\documentclass[lettersize,journal]{IEEEtran}
\usepackage{amsmath,amsfonts}
\usepackage{algorithmic}
\usepackage{algorithm}
\usepackage{array}
\usepackage[caption=false,font=normalsize,labelfont=sf,textfont=sf]{subfig}
\usepackage{textcomp}
\usepackage{stfloats}
\usepackage{url}
\usepackage{verbatim}
\usepackage{graphicx}
\usepackage{cite}
\usepackage{bm}
\usepackage{xcolor}
\usepackage[colorlinks=true, citecolor=blue, linkcolor=blue, urlcolor=blue]{hyperref}

\begin{document}

\title{S4Fusion: Saliency-aware Selective State Space Model for Infrared and Visible Image Fusion}

\author{Haolong Ma, Hui Li*~\thanks{Corresponding author: Hui Li. Email: lihui.cv@jiangnan.edu.cn}~\IEEEmembership{Member,~IEEE}, Chunyang Cheng, Gaoang Wang, \\
Xiaoning Song~\IEEEmembership{Member,~IEEE}, Xiao-Jun Wu~\IEEEmembership{Member,~IEEE}
\thanks{Hui Li, Haolong Ma, Chunyang Cheng, Xiaoning Song, and Xiao-Jun Wu are with Sino-UK Joint Laboratory on Artificial Intelligence of Ministry of Science and Technology, International Joint Laboratory on Artificial Intelligence of Ministry of Education and School of Artificial Intelligence and Computer Science, Jiangnan University, 214122, Wuxi, China.}
\thanks{Gaoang Wang is with the College of Computer Science and Technology, Zhejiang University, Hangzhou, China, 310007.}
\thanks{This work was supported by the National Natural Science Foundation of China (NO.62202205, 62020106012, U1836218), the National Key Research and Development Program of China (2023YFF1105102, 2023YFF1105105) and the Fundamental Research Funds for the Central Universities (JUSRP123030).}
\thanks{The initial version of this article has been posted on arXiv. The link is as follows: https://arxiv.org/abs/2405.20881}
}
 
\markboth{Journal of \LaTeX\ Class Files,~Vol.~14, No.~8, August~2021}%
{Shell \MakeLowercase{\textit{et al.}}: A Sample Article Using IEEEtran.cls for IEEE Journals}


\maketitle

\begin{abstract}

The preservation and the enhancement of complementary features between modalities are crucial for multi-modal image fusion and downstream vision tasks.
However, existing methods are limited to local receptive fields (CNNs) or lack comprehensive utilization of spatial information from both modalities during interaction (transformers), which results in the inability to effectively retain useful information from both modalities in a comparative manner. Consequently, the fused images may exhibit a bias towards one modality, failing to adaptively preserve salient targets from all sources.
Thus, a novel fusion framework (S4Fusion) based on the \textbf{S}aliency-aware \textbf{S}elective \textbf{S}tate \textbf{S}pace is proposed. S4Fusion introduces the Cross-Modal Spatial Awareness Module (CMSA), which is designed to simultaneously capture global spatial information from all input modalities and promote effective cross-modal interaction. This enables a more comprehensive representation of complementary features.
Furthermore, to guide the model in adaptively preserving salient objects, we propose a novel perception-enhanced loss function. This loss aims to enhance the retention of salient features by minimizing ambiguity or uncertainty, as measured at a pre-trained model's decision layer, within the fused images. The code is available at https://github.com/zipper112/S4Fusion.
\end{abstract}


\section{Introduction}

\begin{figure}[ht!]
  \centering
   \includegraphics[width=0.5\textwidth,height=0.3\textwidth]{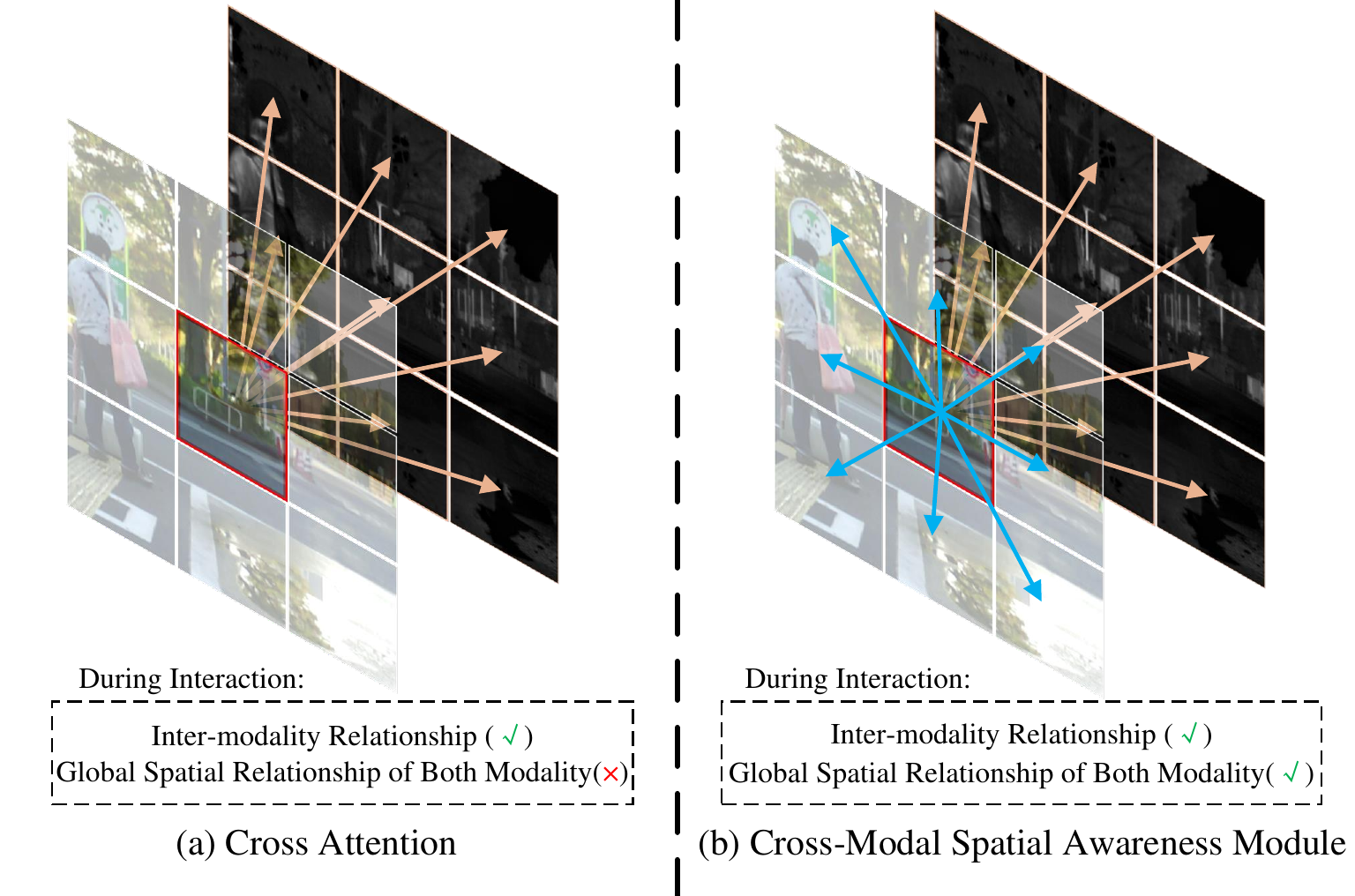}
  \caption{Theoretical interaction region comparison diagram between our SSSM-based CMSA and the mainstream cross attention mechanism. (a) In cross attention, during interaction, only the global spatial information of the opposite modality is captured, while the global spatial information of the same modality is neglected. (b) In contrast, CMSA enables simultaneous attention to both the interaction between modalities and the spatial information of both modalities.}
  \label{fig:first_pic}
\end{figure}
As an image processing technique, Infrared and Visible Image Fusion (IVIF) aims to integrate information from different modalities captured by multiple sensors observing the same scene, producing a comprehensive and informative image \cite{liu2023sgfusion,zhao2020bayesian,ZhaoDIDFuse2020}. 

Visible images typically offer rich textural details and high spatial resolution; however, they are vulnerable to variations in illumination and instances of occlusion, which can result in obscuration or loss of critical scene information \cite{zhang2023visible}. In contrast, infrared images capture thermal radiation and can highlight salient objects even under challenging conditions, such as low illumination. However, they are often susceptible to sensor noise and tend to lack fine textural details \cite{zhang2021image}. Therefore, comprehensive integration of complementary information from these modalities is essential to achieve a more holistic and accurate representation of real-world scenes.

This technique has been widely used in various scenarios such as image restoration \cite{falahkheirkhah2022drb}, UAV
reconnaissance \cite{li2021embedded,sun2021fusion}, image registration \cite{li2023feature}, and can be combined with downstream tasks such as instance segmentation \cite{TANG2023PSFusion}, detection \cite{zhao2023metafusion,wang2023interactively}, and object tracking \cite{peng2023siamese}. 

Originally from control systems, Selective State Space Models (SSSM) are a type of State Space Model (SSM) \cite{gu2021efficiently,gu2021combining}. For the IVIF task, the selective preservation of information from both source images is crucial \cite{liu2023sgfusion}. Previous works overlook the importance of considering the global spatial information of \textbf{both modalities} during the interaction\cite{Tang_2023_DATFuse,liang2022fusion,li2024mambadfuse}. As shown in Fig.\ref{fig:first_pic}, conventional cross attention mechanisms, while effective for inter-modal information exchange, typically direct attention to the spatial features of the complementary modality only. Consequently, the spatial context of the source modality itself is often disregarded during this interaction, hindering a comprehensive perception of complementary details that requires understanding of both self-referenced and cross-referenced spatial contexts. This limitation is particularly critical for tasks like IVIF, where the two modalities contain highly complementary information. 

Thus, a critical need exists for a fusion strategy capable of concurrently processing global spatial information from all input images, while also discerning and integrating the most salient complementary information from the diverse modalities. A novel module called Cross-Modal Spatial Awareness Module (CMSA) is designed. This module can simultaneously facilitate interactions from multi-modalities and within the spatial domains of each modality, thereby enhancing the perception of salient information. This is achieved by unifying modal and spatial interactions through a selective scanning mechanism, enabling the model to retain crucial information from "redundant" source images.

Furthermore, in the real-world scenario, the information from fused images should be determined and structured rather than random noise. Moreover, for models such as ResNet50 \cite{he2016deep} that are pre-trained on classification tasks, they inherently possess the capability to perceive salient targets, which is reflected in their decision layer. Based on this observation, a novel loss function called perception-enhanced loss is designed to drive the CMSA to adaptively highlight salient information. It achieves this by minimizing the uncertainty of a pre-trained model's decision layer on the fused image, thereby enhancing target salience within the image.

With the proposed novel perception-enhanced loss, our method can effectively preserve saliency information from source images.
The contributions of this paper can be summarized as follows:

(\uppercase\expandafter{\romannumeral1}) We design a series of novel modules to capture information from both modalities' own spaces and their interrelations simultaneously, which together constitute the Cross-Modal Spatial Awareness Module. It selectively preserves critical information while eliminating redundant details.

(\uppercase\expandafter{\romannumeral2}) To drive the CMSA to adaptively preserve critical information, a novel loss function is proposed at the decision layer. This function aims to highlight salient information adaptively, resulting in images that better align with visual perception.

(\uppercase\expandafter{\romannumeral3}) Extensive experiments are conducted to validate the effectiveness of our method, including qualitative evaluations, quantitative evaluations, and evaluations involving integration with downstream tasks.

\section{Related Work}
\subsection{Infrared and Visible Image Fusion}
With the rise of deep learning, an increasing number of researchers are adopting deep learning methods to study IVIF tasks \cite{li2018densefuse,li2020nestfuse,li2021rfn}. These methods can generally be classified into three categories: generative models \cite{ma2019fusiongan,ma2020ganmcc}, autoencoder-based models \cite{li2018densefuse,li2020nestfuse}, and end-to-end models \cite{zhang2020rethinking,zhao2023cddfuse}. Generative models often suffer from unstable training, autoencoder-based methods offer more stable training but still rely on hand-crafted fusion strategies \cite{zhang2021image}. End-to-end methods, on the other hand, often design loss functions at the shallow pixel level, which can still be considered as hand-crafted fusion rules, limiting the network's adaptability. However, because of their training stability and simplicity of implementation, the end-to-end method has become the dominant paradigm. To obtain higher-quality fused images, existing end-to-end methods primarily focus on refining improvements in network architecture and loss functions, particularly perceptual loss.

\subsubsection{Perceptual Loss}
Mainstream end-to-end models integrate information from two source images by designing perceptual losses that leverage pre-trained networks \cite{liu2023sgfusion,liang2022fusion,tang2022image}. For example, U2Fusion \cite{xu2022u2fusion} utilizes intermediate features of a pretrained VGG16 as part of the loss function design. Recent advancements, such as LDFusion \cite{wang2024infrared}, have extended this by incorporating intermediate features from vision-language models like CLIP \cite{radford2021learning}. Critically, all these methods exclusively consider the feature layers of pre-trained networks for perceptual loss design, overlooking the potential of the decision layer. The decision layers of pre-trained networks, such as the final classification layer in ResNet, are widely recognized for encoding representations strongly correlated with salient image content and semantic meaning \cite{arandjelovic2017look,caron2021emerging}. Consequently, we propose a novel perception-enhanced loss formulated at the decision layer of a pre-trained model, designed to encourage the fusion network to generate images where salient targets are more distinct.

\subsubsection{Architecture}
Mainstream end-to-end image fusion models primarily leverage either Convolutional Neural Networks (CNNs) \cite{xu2022u2fusion,liang2022fusion,li2018densefuse} or Transformers \cite{zhao2023equivariant,liu2022target,li2025maefuse}. While CNNs exhibit strong inductive biases for image data, their constrained local receptive fields inherently limit their capacity for global information extraction \cite{ding2022scaling}. Conversely, Transformer-based models have recently gained prominence in image fusion due to their superior global spatial modeling capabilities. For example, EMMA \cite{zhao2023equivariant} employs Restormer's efficient channel attention for hierarchical feature extraction and modal interaction, while CrossFuse \cite{li2024crossfuse} integrates both spatial and channel attention for feature extraction alongside cross-attention for modal interaction. However, a significant drawback of many Transformer-based models is their quadratic computational complexity with respect to input sequence length, often leading to considerable computational overhead.

The core task of image fusion—preserving critical information from both source images—requires selective retention and filtering. Many existing Transformer-based interaction modules are not explicitly optimized for filtering redundant information and may not sufficiently attend to the spatial details within each modality concurrently during the interaction process. Selective State Space Models, particularly Mamba \cite{gu2023mamba}, with their robust global modeling and adaptive selection mechanisms, offer a promising solution to these limitations.

\subsection{Selective State Space Model}
Recently, State Space Models have attracted increasing interest from researchers across various fields, demonstrating significant potential in Natural Language Processing (NLP) and Computer Vision (CV) \cite{lu2024structured,ruan2024vm,nguyen2022s4nd}. Early works like LSSL \cite{gu2021combining}, when combined with HiPPO \cite{gu2020hippo} initialization, exhibited powerful capabilities in capturing long-range dependencies. With the emergence of Mamba \cite{tridao2024mamba,gu2023mamba}, the Selective State Space Model was introduced and achieved remarkable success in NLP. The introduction of VMamba \cite{liu2024vmamba} brought Mamba into the field of CV. Its powerful global receptive field, rapid inference speed, and lack of need for positional embedding have inspired many related works in CV.

In the image fusion task, some works start to combine with Mamba 
\cite{li2024mambadfuse,cao2024novel}. For example, MambaDFuse \cite{li2024mambadfuse} designed a cross-scanning mechanism, inspired by cross-attention. However, these methods focus primarily on mimicking Transformer-style inter-modal interactions, overlooking the importance of simultaneously considering the spatial information of both modal images. To address this, we propose a new module called CMSA. It unifies spatial and modal interactions by leveraging the selective scanning capability of SSSM, thereby promoting the retention of critical information.

\subsection{Comparison with existing fusion modules}
The proposed CMSA integrates both intra-modal spatial interactions and inter-modal interactions within the selective scanning mechanism characteristic of SSSMs. This allows for a comprehensive consideration of information retention from both modalities and the filtering of redundant information. Relative to conventional Transformer-based fusion modules, CMSA is designed to concurrently process spatial information from both modalities alongside their interactions. Furthermore, it incorporates an inherent structural capacity for adaptive information retention while circumventing the quadratic computational complexity associated with standard attention mechanisms. Furthermore, relative to existing Mamba-based methods, our approach is the first to apply SSSM to unify modal and spatial interactions, rather than merely mimicking Transformer-style interactions within the Mamba structure.

\begin{figure*}[tb]
  \centering
   \includegraphics[width=1\textwidth,height=0.35\textwidth]{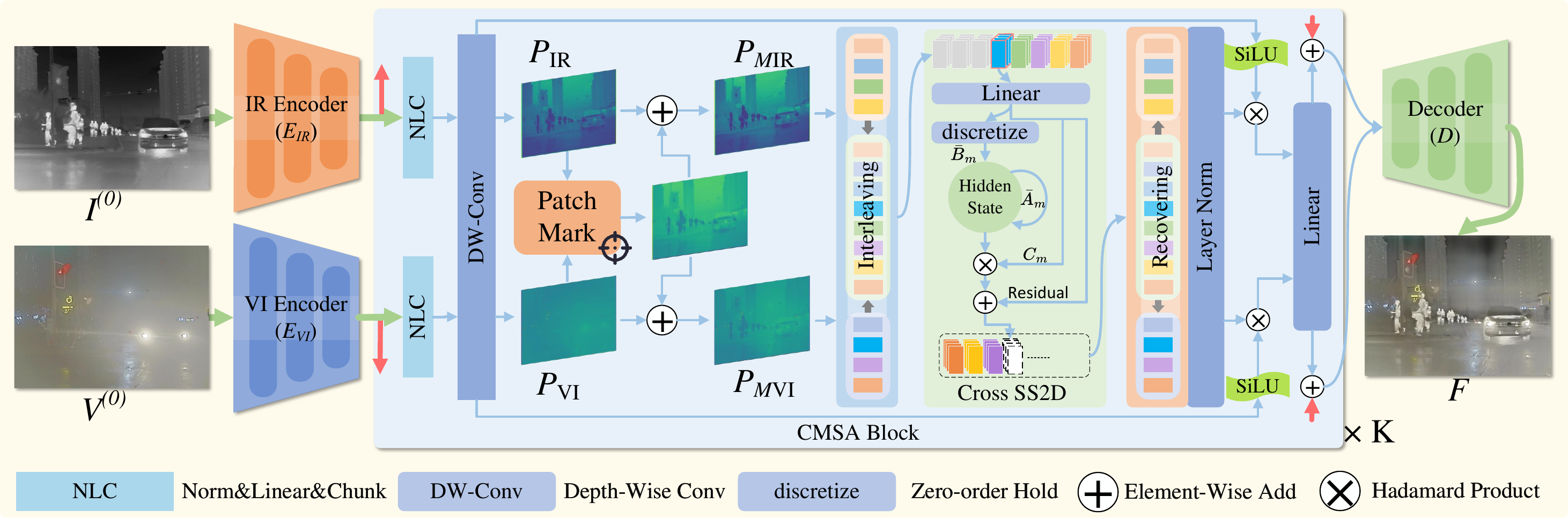}
  \caption{Architectural diagram of S4Fusion. It consists of two encoders, a decoder, and our novel CMSA block. The illustrative diagram of the CMSA block is suitable for use with the output of any encoder layer.}
  \label{fig:overall}
\end{figure*}

\section{Proposed Method}
This section presents the details of our proposed method. As shown in Fig.\ref{fig:overall}, our model adopts a multi-scale structure to process source images. It can be broadly divided into three main components: the encoder, the decoder, and the fusion module. For simplicity, we assume that both the encoder and decoder contain \( N \) layers.

\subsection{Preliminary}
\label{sec:1}
\subsubsection{SSM}
\label{pre}
The SSM initially served as a time-series processing model, mapping input sequences $x(t)$ to output sequences $y(t)$ through hidden states $h(t)$ to memorize and capture temporal relationships. Here, $t$ represents a continuous variable denoting time. Formally, SSM can be expressed by the following two equations:
\begin{gather}
h'(t) = A_mh(t) + B_mx(t)\\
y(t) = C_mh(t)
\end{gather}
where $A_m$, $B_m$, and $C_m$ are parameters. $A_m$ controls the update of hidden states, while $B_m$ and $C_m$ govern how input data flows into and out of hidden states.
Subsequently, to better handle discrete data in real-world scenarios (such as text), the time step parameter $\Delta$ is introduced to control the time interval. Through zero-order hold \cite{gu2021efficiently}, SSM is further discretized into the following form:
\begin{gather}
h_t = \bar{A}_mh_{t-1} + \bar{B}_mx_t\\
y_t = C_mh_t
\end{gather}
where $\bar{A}_m$ and $\bar{B}_m$ are counterparts of $A_m$ and $B_m$ after zero-order hold, with $\bar{A}_m = \exp(\Delta A_m)$ and $\bar{B}_m = \Delta B_m$. After discretization, the subscript $t$ represents the $t$-th input in the sequence, while $\Delta$ records the distance between adjacent elements in the sequence.

\subsubsection{Mamba and VMamba}
Mamba \cite{tridao2024mamba} transforms the original parameters \( A_m, B_m \), and \( C_m \) into context-dependent parameters, making SSM selective to inputs . Specifically, at time \( t \), the computation process of SSM from input to output utilizes parameters \( A_m \), \( B_m \), and \( C_m \) derived from the input \( x_t \).

\begin{figure}[tb]
  \centering
   \includegraphics[width=0.45\textwidth,height=0.23\textwidth]{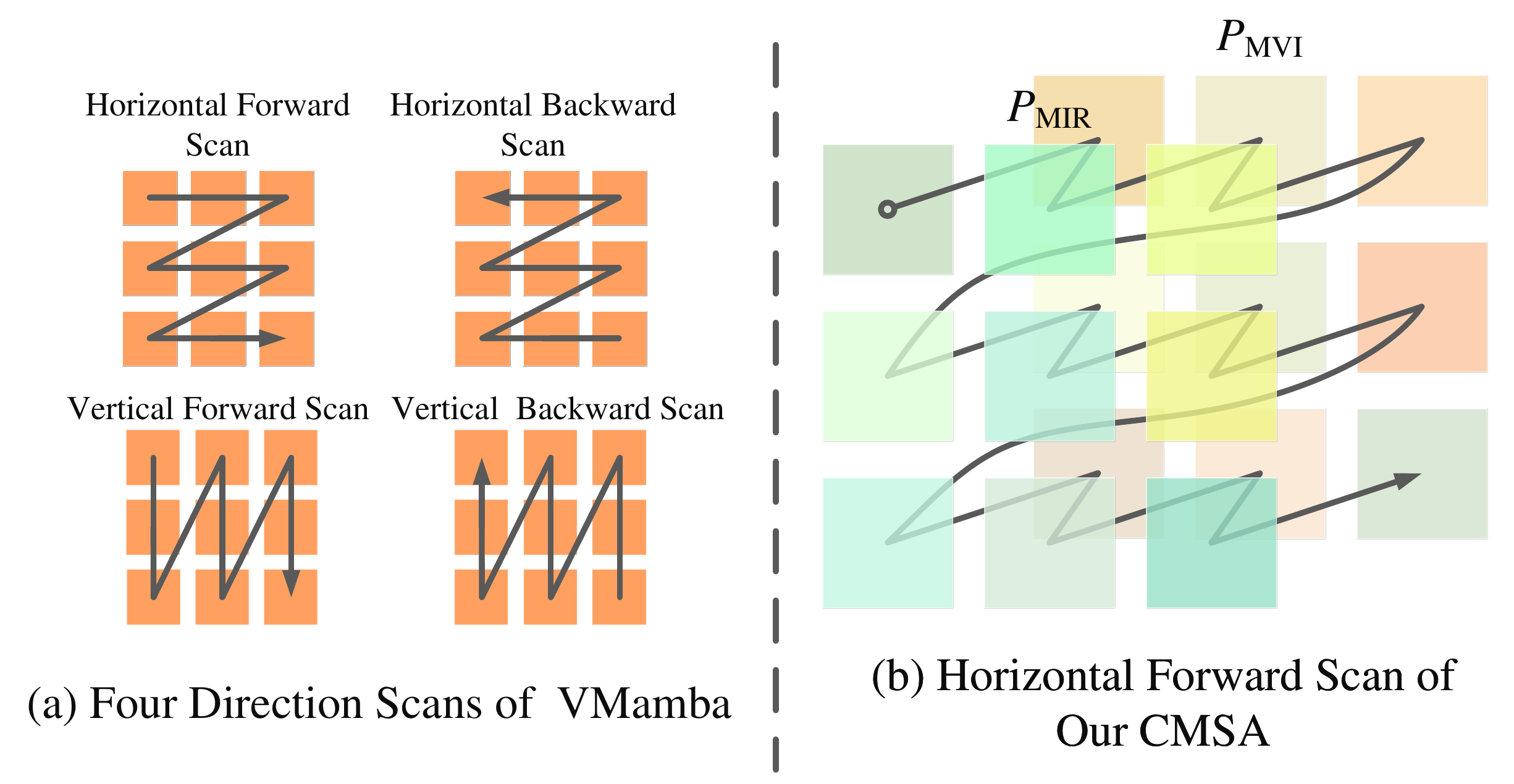}
  \caption{(a) shows the four scanning trajectories of SS2D as described in the VMamba. (b) illustrates the Cross SS2D scanning trajectory in the Horizontal Forward direction. This corresponds to the improvement of the Horizontal Forward direction in SS2D within our Cross SS2D, and the other three directions are similar.}
  \label{fig:cross_scan_merge}
\end{figure}

Subsequently, VMamba \cite{ruan2024vm} introduces Mamba into the realm of CV for the first time and proposes the SS2D module to address the inability of Mamba to handle image data. As shown in Fig.\ref{fig:cross_scan_merge}(a), VMamba non-overlappingly divides the image into patches and flattens them into sequences in four different directions. These sequences are then individually inputted into Mamba blocks for computation and finally aggregated to extract features.
\subsection{Encoder and Decoder}
Comprising an Overlap Patch Embedding (OPE) and multiple encoder layers, the encoder processes input images. The decoder, on the other hand, utilizes a layer operation and several layers to stitch the final output features back into an image. The OPE's role is to encode inputs with an overlapping strategy. This strategy helps prevent artifacts in the fusion results by dividing the input image into overlapping patches where adjacent patches share a portion of their pixels.

We utilize the VSS block \cite{ruan2024vm} in both the encoder and decoder. In the encoder, it extracts basic features from individual modality images, while in the decoder, it integrates features of fused images across different scales. Its design, similar to the Mamba block, enables global information modeling.

Let \( I \) and \( V \) denote two source images. After applying OPE, we obtain \(I^{(0)},V^{(0)}\in \mathbb{R}^{H\times W\times C}\), where \(C\) is the number of channels and \(H \times W\) is the size of the feature map. Specifically, for the \(l\)-th layer of the encoder:
\begin{equation}
\begin{gathered}
 I^{(l)}=Down^{(l)}(E_{\mathrm{IR}}^{(l)}(I^{(l-1)})) \\
 V^{(l)}=Down^{(l)}(E_{\mathrm{VI}}^{(l)}(V^{(l-1)})), \\ 
 s.t. \  l \in\{1,..,N\}
\end{gathered}
\end{equation}
where \( E_{\mathrm{IR}}^{(l)}(\cdot) \) and \( E_{\mathrm{VI}}^{(l)}(\cdot) \) represent the encoder for the infrared and visible modalities at the \( l \)-th layer. \( Down^{(l)}(\cdot) \) denotes Patch Merging or \( Identity \) operation and \( Down^{(l)}(\cdot)=Identity(\cdot) \) only when \( l=N \).

For the decoder at the \( l \)-th layer, it takes the fused feature \( R^{(l)} \) and the output from the previous scale \( F^{(l)} \) as input to reconstruct the fused features at corresponding scale. The formulas are given as follows:
\begin{equation}
\begin{gathered}
F^{(l-1)}=D^{(l)}(UP^{(l)}(F^{(l)})+R^{(l)}), \\  
s.t.\ F^{(N)}=I^{(N)}+V^{(N)}, l\in\{1,2,\cdots,N\} 
\end{gathered}
\end{equation}
where \( UP^{(l)} \) represents Patch Expanding or \( Identity \) operation. Similar to \( Down^{(l)}(\cdot) \), \( UP^{(N)}(\cdot)=Identity(\cdot) \).

Finally, the output \( F^{(0)} \) undergoes the fold operation to remove overlaps and is linearly mapped to integrate channels, resulting in the final fused image \( F \).

\subsection{Cross-Modal Spatial Awareness Module}
The CMSA is proposed as the fusion module to interact with features from the infrared and visible modalities at various scales, thereby efficiently perceiving salient information. Specifically, CMSA consists of \( K \) CMSA Blocks (CMSAB).

For ease of discussion, in this section, we will use the \( k \)-th CMSAB at the \( l \)-th layer as an example. Furthermore, \(V^{(l,0)}=V^{(l)}\) and \(I^{(l,0)}=I^{(l)}, k \in \{1,2,\cdots,K\}, l \in \{1,2..,N\}\).

As depicted in Fig.\ref{fig:overall} (CMSA Block), features extracted from two modalities are first fed into two branches separately using a layer norm, a linear mapping and the channel-wise chunk operation (NLC). One branch directly passes through a SiLU activation function and is then used for gated output, while the other passes through a shared \(3 \times 3\) depth-wise convolution to extract local features, denoted as \( P^{(l,k)}_{\mathrm{IR}} \) and \( P^{(l,k)}_{\mathrm{VI}} \), respectively. 
 
The above two features are then utilized for feature interaction, which consists of four steps: \textbf{Patch Mark}, \textbf{Interleaving}, \textbf{Cross SS2D}, and \textbf{Recovering}. Subsequently, the two obtained outputs go through a layer norm, a linear layer and a residual connection, yielding the outputs \( I^{(l,k+1)} \) and \( V^{(l,k+1)} \) of the \(k\)-th CMSAB. 
 
At the end of the \( K \)-th CMSAB, the two deeply interacted features are element-wise summed: \( R^{(l)}=I^{(l,K)}+V^{(l,K)} \), resulting in the output \( R^{(l)} \) of the $l$-th CMSA.

\begin{figure}[tb]
  \centering
   \includegraphics[width=0.30\textwidth,height=0.23\textwidth]{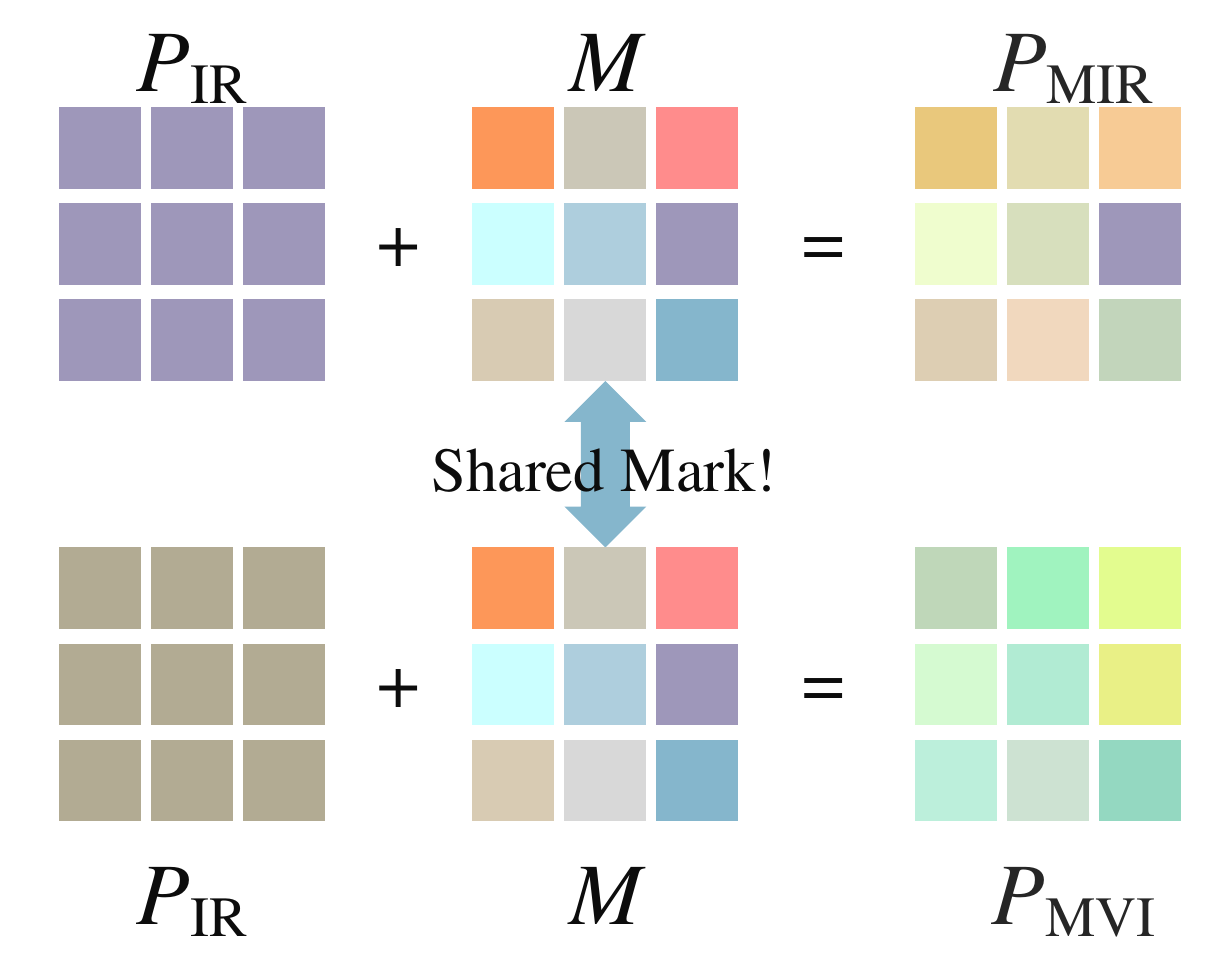}
  \caption{Schematic diagram of Patch Mark. Patches at the same spatial location across different modalities are grouped by adding a mark, similar to a positional encoding.}
  \label{fig:mark}
\end{figure}
\subsubsection{\textbf{Patch Mark}}
To efficiently process multi-modal interaction, we propose Patch Mark operation, which is inspired by Positional Encoding. It is used to group patches at corresponding positions across two modalities.

As shown in Fig.\ref{fig:mark} and Fig.\ref{fig:overall}, Patch Mark concatenates the two inputs along the channel dimension and then applies a linear layer to generate a mark for each patch position. Consequently, patches from the same position in each modality are labeled with the same mark, resulting in \( P^{(l,k)}_{\mathrm{MIF}} \) and \( P^{(l,k)}_{\mathrm{MVF}} \):
\begin{equation}
M^{(l,k)}=linear^{(l,k)}([P^{(l,k)}_{\mathrm{IR}};P^{(l,k)}_{\mathrm{VI}}])
\end{equation}
\begin{equation}
P^{(l,k)}_{\mathrm{MIR}}=P^{(l,k)}_{\mathrm{IR}}+M^{(l,k)},P^{(l,k)}_{\mathrm{MVI}}=P^{(l,k)}_{\mathrm{VI}}+M^{(l,k)}
\end{equation}
where \( [\cdot;\cdot] \) denotes concatenation along with the channel dimension, and \( M^{(l,k)} \) represents the mark at \(l\)-th layer and \(k\)-th CMSAB. Patch Mark effectively groups patches from the same positions in both modalities, enabling a clear distinction between inter-modal interactions and spatial interactions.

\subsubsection{\textbf{Interleaving}}
\begin{figure*}[tb]
  \centering
   \includegraphics[width=1\textwidth,height=0.35\textwidth]{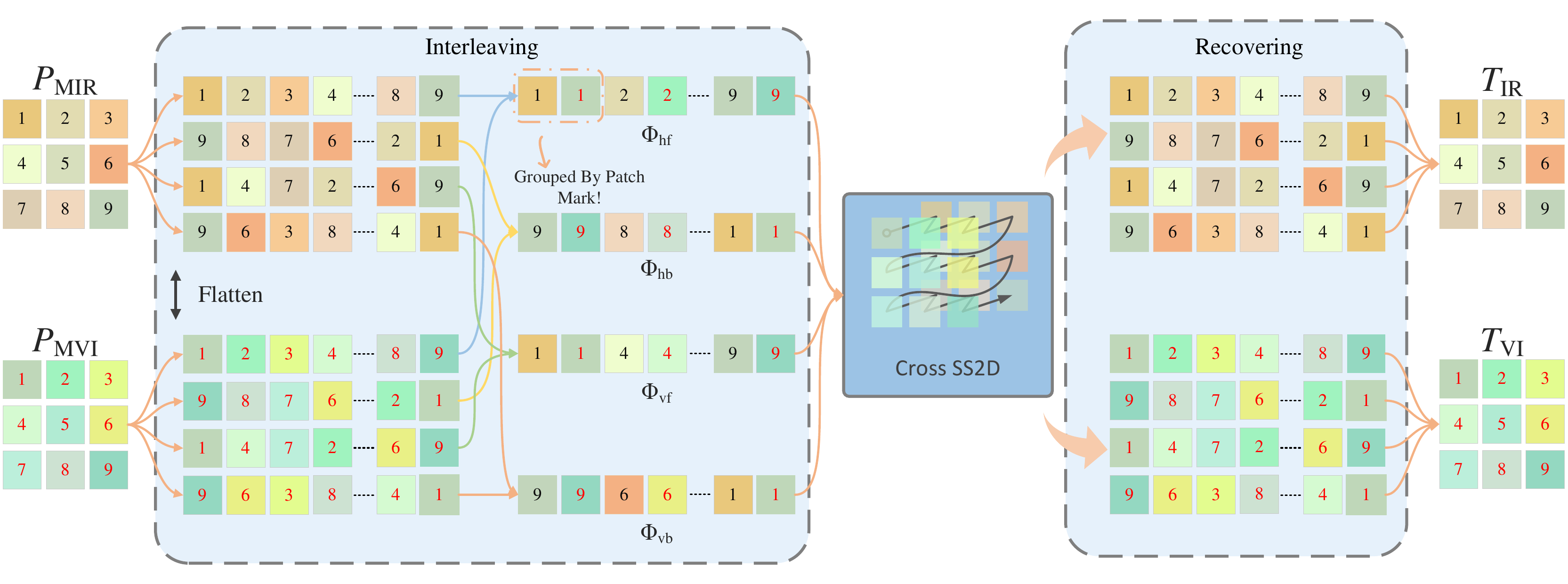}
  \caption{Demonstration of the Interleaving and Recovering processes. The numbers in different colors represent patches from different modalities, and identical numbers denote patches from the same position in the feature maps.}
  \label{fig:recov_inter}
\end{figure*}

The interleaving operation flattens the patches extracted from two modalities into sequences in four directions. Subsequently, sequences from the same unfolding direction in the two modalities are merged alternately into a new sequence. As depicted in Fig.\ref{fig:recov_inter}, the features of two modalities are initially unfolded into eight sequences based on combinations of vertical ($v$), horizontal ($h$), forward ($f$), and backward ($b$) directions. 

Then, sequences with the same flattening direction are interleaved to form four long sequences: 
\begin{equation}
\Phi^{(l,k)}_{hf},\Phi^{(l,k)}_{hb},\Phi^{(l,k)}_{vf},\Phi^{(l,k)}_{vb}=inter(P^{(l,k)}_{\mathrm{MIR}}, P^{(l,k)}_{\mathrm{MVI}})
\end{equation}
where \(inter\) denotes the interleaving operation. Subscripts \(hf\) and \(vb\) represent Horizontal Forward and Vertical Backward flattening directions, respectively, and similarly for others. 

Taking \(\Phi^{(l,k)}_{hf}\) in Fig.\ref{fig:recov_inter} as an example, due to Patch Mark grouping patches, patches with the same number can be considered as one entirety. Each group of patches is considered as a single entity, representing the spatial dimension from left to right across the entire sequence.

\subsubsection{\textbf{Cross SS2D}}
We design Cross SS2D to achieve inter-modal interaction that simultaneously attends to spatial information from two modalities. Cross SS2D accomplishes this by scanning interleaved sequences in four directions. 

The core idea of Cross SS2D is that \textbf{Modality-specific parameters are used to control complementary information between modalities, while the shared hidden state represents common information across modalities}. This means that for the parameters \(\Delta\), \(A_m\), and \(B_m\), the infrared and visible modalities are private. By using modality-specific private parameters, each modality can selectively compress and store information related to other modalities into the hidden state during the scanning process. Conversely, each modality can selectively extract information related to itself from the hidden state as output, thus completing the interaction. 

As shown in Fig.\ref{fig:cross_scan_merge}(b), Cross SS2D continuously scans between two modalities to achieve inter-modal interaction. Thanks to the grouping effect of Patch Mark, patches at the same position in different modalities are grouped as one entirety. Consequently, when performing inter-modal interactions, \textbf{Cross SS2D facilitates spatial interactions between the two modalities at the group level, enabling the consideration of saliency information from a global spatial perspective.} Specifically, this process can be succinctly described as follows: 
\begin{equation}
F^{(l,k)}_{d}=CSS2D(\Phi^{(l,k)}_{d}), (d\in \{hf,hb,vf,vb\})    
\end{equation}
where $CSS2D$ denotes Cross SS2D. 

Algorithm \ref{alg1} describes the Cross SS2D algorithm in a manner similar to Mamba. Assuming the input sequence length is $2L$, where $L$ is the number of patches in the feature map, $H$ and $B$ represent the size of the hidden state in SSM and the batch size, respectively. $S_{(\cdot)}$ denotes a linear layer, where the first letter in the subscript indicates the perceived parameter(\textit{e.g.} \(B_m\) and \(C_m\)), and the second letter denotes the modality (\textit{i.e.} I and V). \(\tau_{\Delta}\) represents softplus function. 

Since the sequences of the two modalities are interleaved, we must also interleave the parameters to ensure that each modality has independent control over its input and output.
$Inter(\cdot,\cdot)$ represents the interleaving arrangement of linear layers from different modalities. This means that parameters $\Delta$, $B_m$ and $C_m$ have different perceptions for different modalities. \(discretize(\cdot, \cdot, \cdot)\) represents the process of discretization. The discretization formulas for $\overline{\textbf{A}}$ and $\overline{\textbf{B}}$ are as follows:
\begin{equation}
 \overline{\textbf{A}} = \exp(\Delta \textbf{A}), \overline{\textbf{B}}  = (\Delta \textbf{A})^{-1} (\exp (\Delta \textbf{A}) - \textbf{I}) \cdot \Delta \textbf{B}
\end{equation}
where $\Delta$, $\textbf{A}$, and $\textbf{B}$ are, respectively, the three inputs for discretization.
We provide a more detailed interpretation of these parameters in Sec. \ref{Interpretation}.

\begin{algorithm} 
	\caption{Cross  SS2D} 
	\label{alg1}
        \renewcommand{\algorithmicrequire}{\textbf{Input:}}
        \renewcommand{\algorithmicensure}{\textbf{Output:}}
	\begin{algorithmic}
		\REQUIRE $\textbf{x}$: (B, 2L, C) 
		\ENSURE $\textbf{y}$: (B, 2L, C) 
		\STATE $\bm{A_m}: (B, L, H) \gets Param$ 
            \STATE $\bm{B_m}: (B, 2L, H) \gets Inter(S_{\mathrm{B_mI}},S_{\mathrm{B_mV}})(\textbf{x})$
            \STATE $\bm{C_m}: (B, 2L, H) \gets Inter(S_{\mathrm{C_mI}},S_{\mathrm{C_mV}})(\textbf{x})$ 
            \STATE $\Delta : (B, 2L, H) \gets \tau_{\Delta}(Param + Inter(S_{\mathrm{\Delta I}},S_{\mathrm{\Delta V}})(\textbf{x}))$ 
            \STATE $\overline{\bm{A_m}}, \overline{\bm{B_m}}: (B, 2L, C, H) \gets discretize(\Delta, \bm{A_m}, \bm{B_m})$
            \STATE $\textbf{y} \gets SSM(\overline{\bm{A_m}}, \overline{\bm{B_m}}, \bm{C_m})(\textbf{x})$
            \STATE \textbf{return y}
	\end{algorithmic} 
\end{algorithm}

\subsubsection{\textbf{Recovering}}
Recovering is responsible for disassembling the features after Cross SS2D interaction and restoring them to their original feature map shapes. This process is illustrated in Fig.\ref{fig:recov_inter}. 

Initially, the features are disassembled into sequences organized in their original four directions. Patches at corresponding positions within each modality are added together and combined to reconstruct the original shape of the feature map:
\begin{equation}T_{\mathrm{IR}}^{(l,k)},T_{\mathrm{VI}}^{(l,k)}=Recover(F^{(l,k)}_{hf},F^{(l,k)}_{hb},F^{(l,k)}_{vf},F^{(l,k)}_{vb})
\end{equation}

The recovered features, as shown in Fig.\ref{fig:overall}, are passed through layer normalization, a gating mechanism, and residual connections to obtain the output $V^{(l,k+1)},I^{(l,k+1)}$ of the $k$-th CMSAB layer.

\subsection{Loss Function}
The loss function comprises four components: a L1 loss \(\mathcal{L}_{\mathrm{L1}}\), a structural similarity loss \(\mathcal{L}_{\mathrm{ssim}}\), a gradient loss \(\mathcal{L}_{\mathrm{grad}}\), and the proposed perception-enhanced loss \(\mathcal{L}_{\mathrm{per}}\). The total loss \(\mathcal{L}_{\mathrm{total}}\) is given as follows,
\begin{equation}
\mathcal{L}_{\mathrm{total}}=\mathcal{L}_{\mathrm{per}}+\alpha_1 \mathcal{L}_{\mathrm{L1}}+\alpha_2 \mathcal{L}_{\mathrm{ssim}}+\alpha_3 \mathcal{L}_{\mathrm{grad}}
\end{equation}
where \(\alpha_1\), \(\alpha_2\), and \(\alpha_3\) are hyperparameters used to balance each item.

\(\mathcal{L}_{\mathrm{ssim}}\) and \(\mathcal{L}_{\mathrm{L1}}\) are inspired by U2Fusion \cite{xu2022u2fusion}. We use a pre-trained ResNet50 to measure the amount of information of the two original images, obtaining estimates \(\omega_1\) and \(\omega_2\) to compute \(\mathcal{L}_{\mathrm{L1}}\) and \(\mathcal{L}_{\mathrm{ssim}}\).

To approximate human visual perception and enhance saliency, the perception-enhanced loss $\mathcal{L}_{\mathrm{per}}$ is designed at the decision layer of the pre-trained ResNet50 \cite{he2016deep}. For the fusion result $F$, if its salient targets are more prominent, it should contain more definitive information rather than noise.

We assume \(y\) as the probability distribution of ResNet50's estimates for \(F\) over 1000 categories. The entropy of this distribution estimates the certainty, reflecting the saliency target perception: 
\begin{equation}
\begin{gathered}
\mathcal{L}_{\mathrm{per}}=-\sum\limits_{i=1}^{1000}y_ilog(y_i) \\
where \ \ y=\mathrm{softmax}(ResNet50(F))
\end{gathered}
\end{equation}

\(\mathcal{L}_{\mathrm{per}}\) minimizes this uncertainty to highlight salient objects in the two original images. However, focusing solely on salient targets may lead to loss of texture details and degrade image quality. Moreover, because \(\mathcal{L}_{\mathrm{per}}\) emphasizes using the pre-trained network to strengthen semantic regions in the image, this can lead to a reduced focus on texture details, further degrading image quality. Thus, we introduce \(\mathcal{L}_{\mathrm{grad}}\) to preserve texture details while restraining the perception-enhanced loss: \(\mathcal{L}_{\mathrm{grad}}=|||\nabla F|-\max(|\nabla I|,|\nabla V|)||_1\), where \(\nabla\) represents the Sobel operator used to compute the edge texture information of the image.

\begin{figure*}[!tb]
  \centering
   \includegraphics[width=1\textwidth,height=0.29\textwidth]{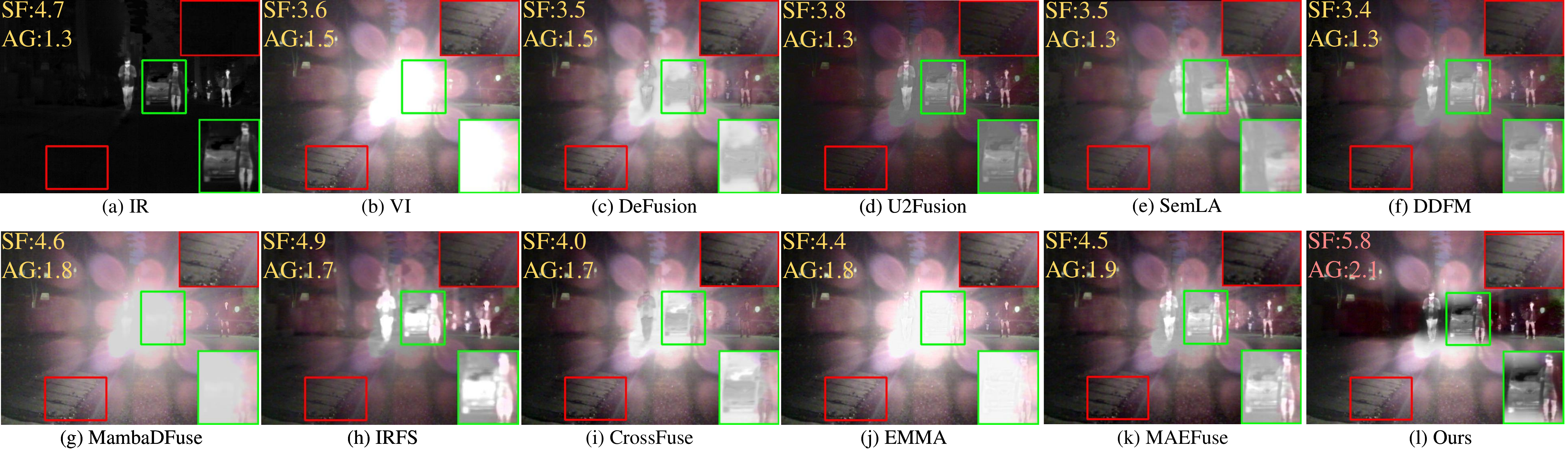}
  \caption{Qualitative comparison of the original images and the results of all methods on the image ``00037N" in the MSRS \cite{tang2022image} dataset.}
  \label{fig:msrs}
\end{figure*}

\begin{figure*}[!tb]
  \centering
   \includegraphics[width=1\textwidth,height=0.29\textwidth]{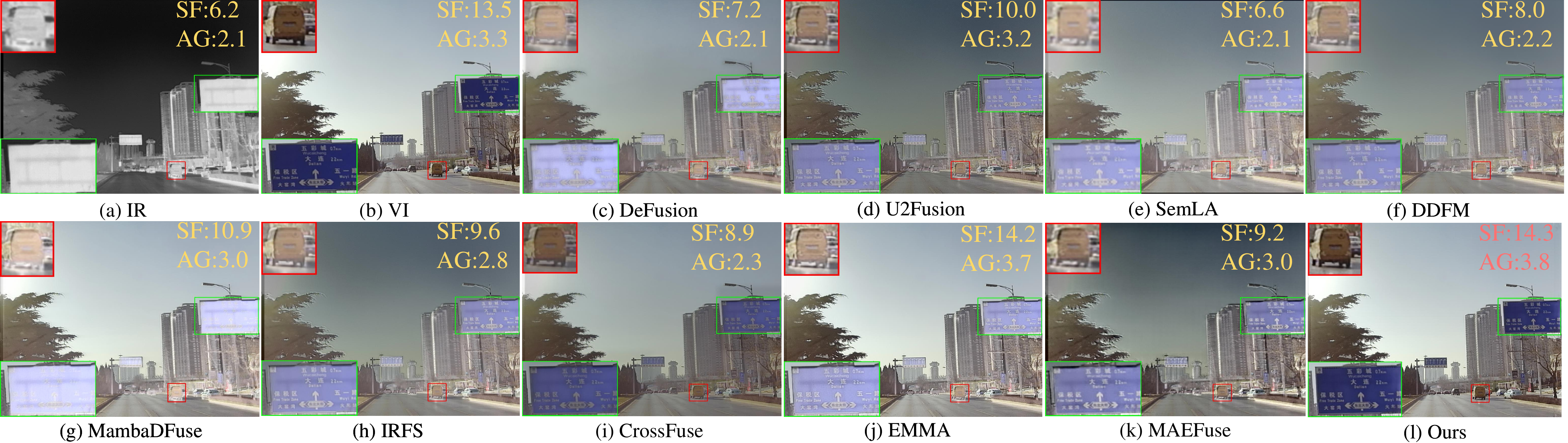}
  \caption{Qualitative comparison of the original images and the results of all methods on the image ``02119" in the M3FD \cite{liu2022target} dataset.}
  \label{fig:m3fd1}
\end{figure*}

\begin{figure*}[!tb]
  \centering
   \includegraphics[width=1\textwidth,height=0.29\textwidth]{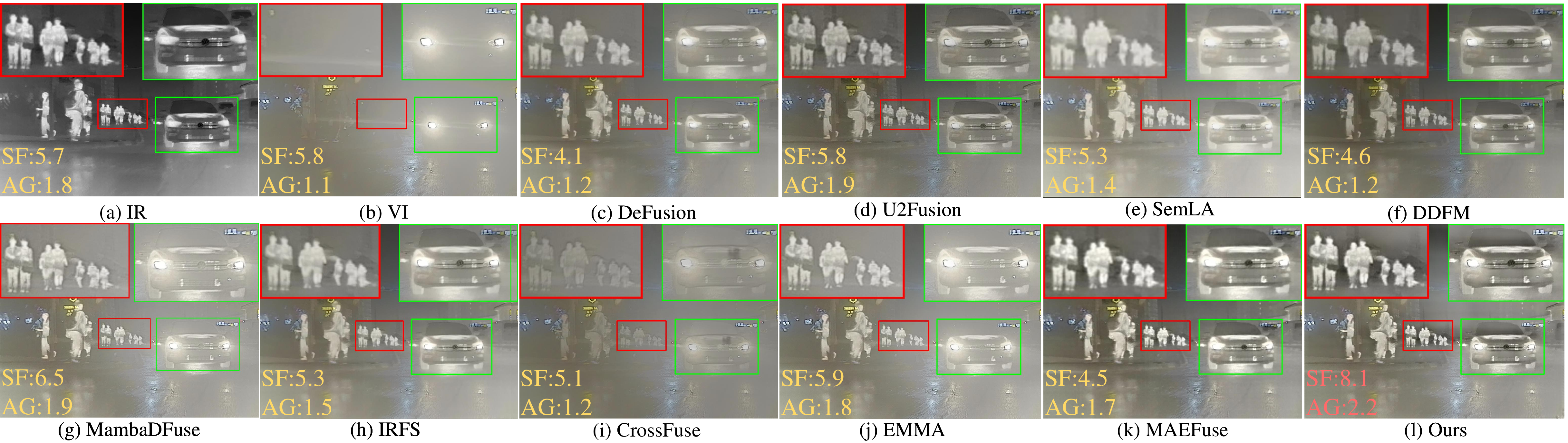}
  \caption{Qualitative comparison of the original images and the results of all methods on the image ``00386" in the M3FD \cite{liu2022target} dataset.}
  \label{fig:m3fd2}
\end{figure*}

\begin{figure*}[!tb]
  \centering
   \includegraphics[width=1\textwidth,height=0.29\textwidth]{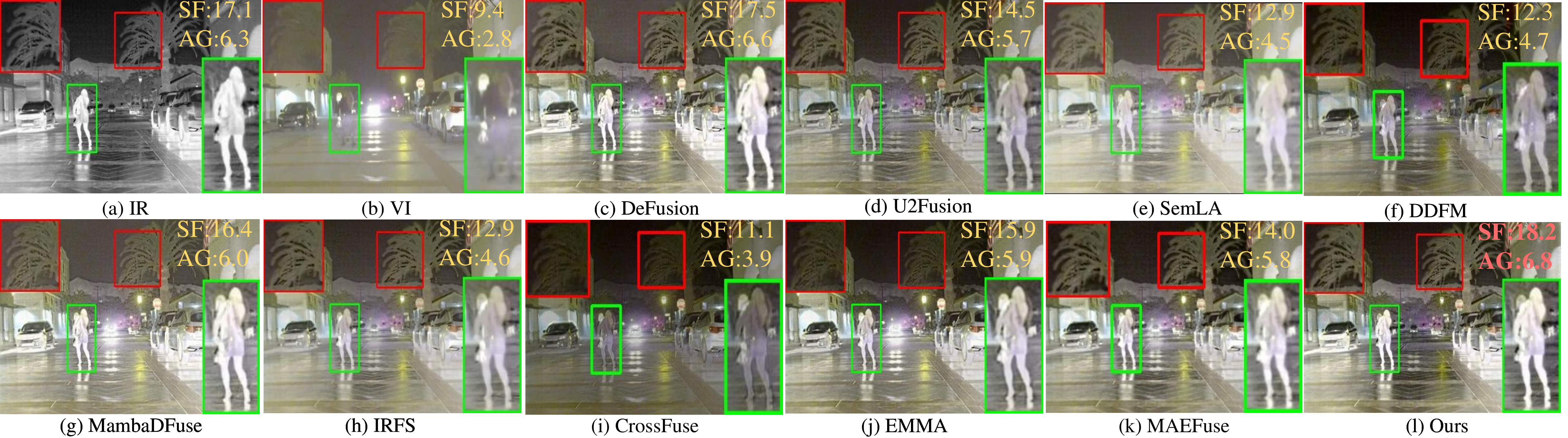}
  \caption{Qualitative comparison of the original images and the results of all methods on the image ``00386" in the RoadScene \cite{xu2022u2fusion} dataset.}
  \label{fig:roadscene}
\end{figure*}

\begin{figure*}[!tb]
  \centering
   \includegraphics[width=1\textwidth,height=0.30\textwidth]{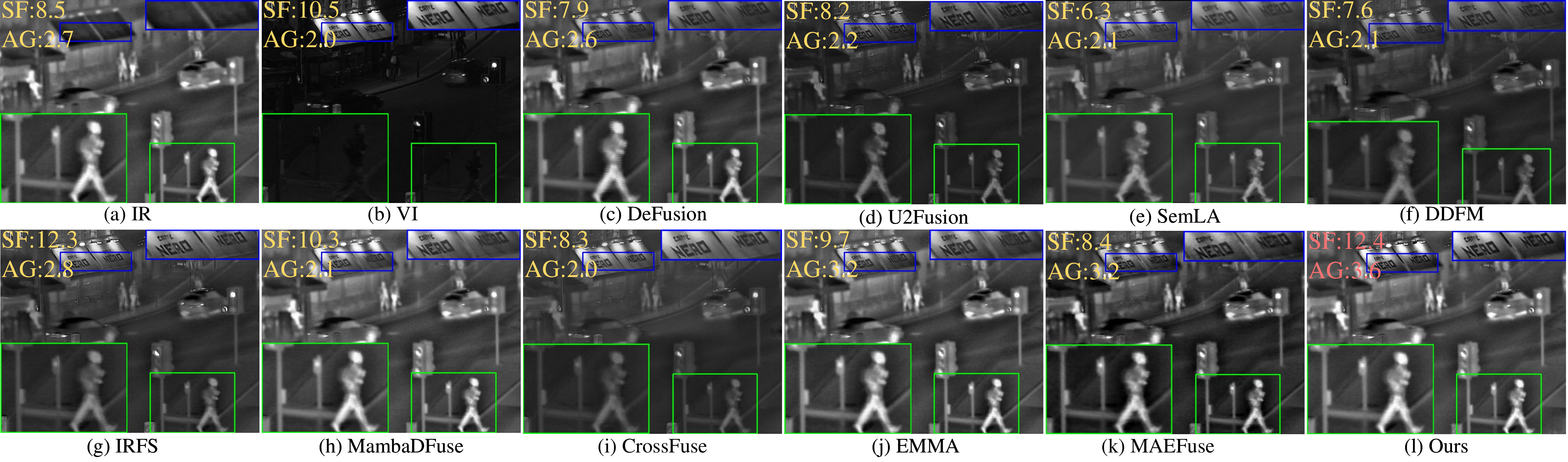}
  \caption{Qualitative comparison of the original images and the results of all methods on the image ``Road" in the TNO dataset.}
  \label{fig:TNO}
\end{figure*}

\begin{figure*}[!tb]
  \centering
   \includegraphics[width=1\textwidth,height=0.69\textwidth]{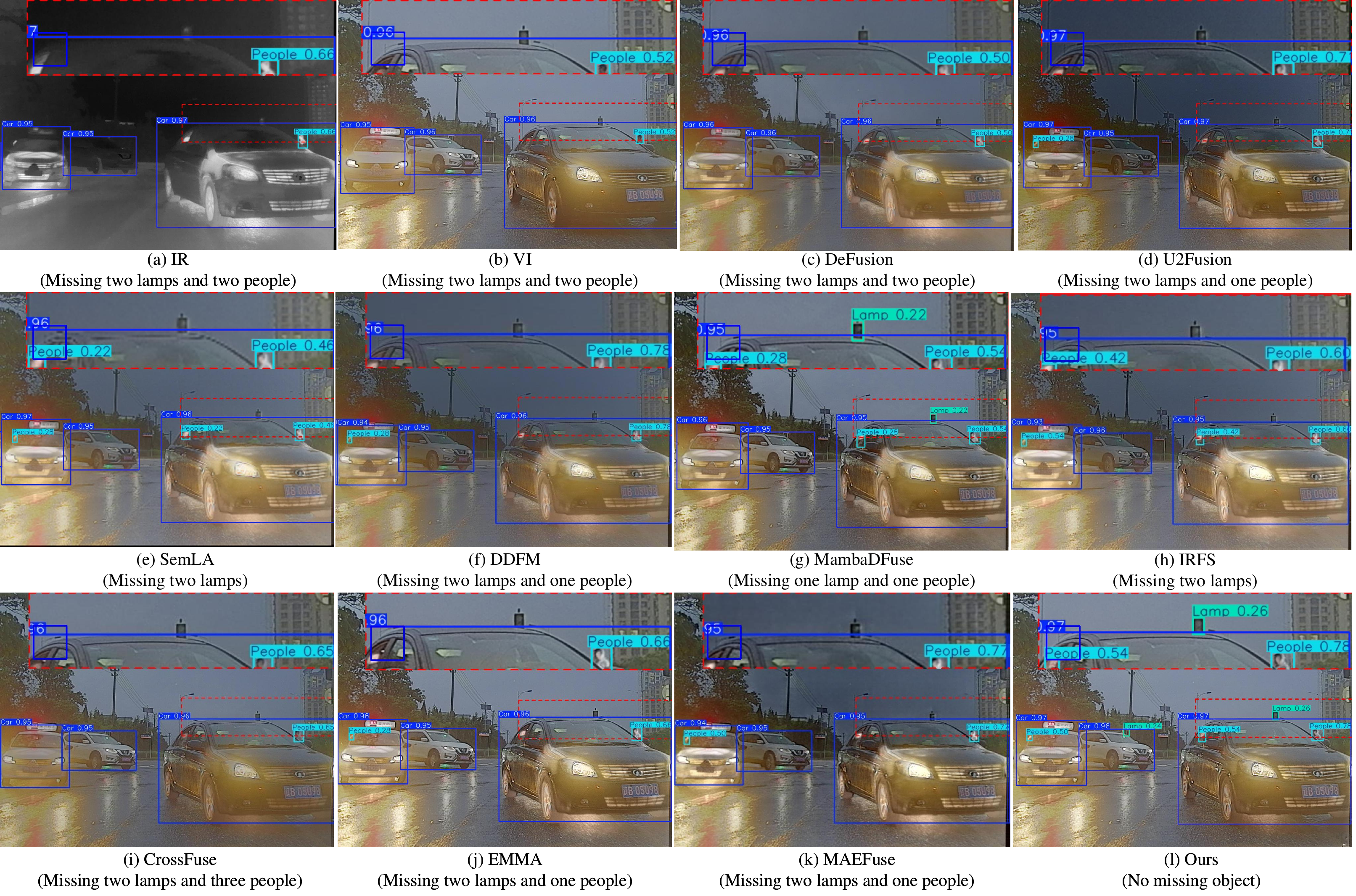}
  \caption{Qualitative comparison of Multi-modal Object Detection on the image ``00215" in the M3FD \cite{liu2022target} dataset.}
  \label{fig:detect}
\end{figure*}

\begin{figure*}[!tb]
  \centering
   \includegraphics[width=1\textwidth,height=0.43\textwidth]{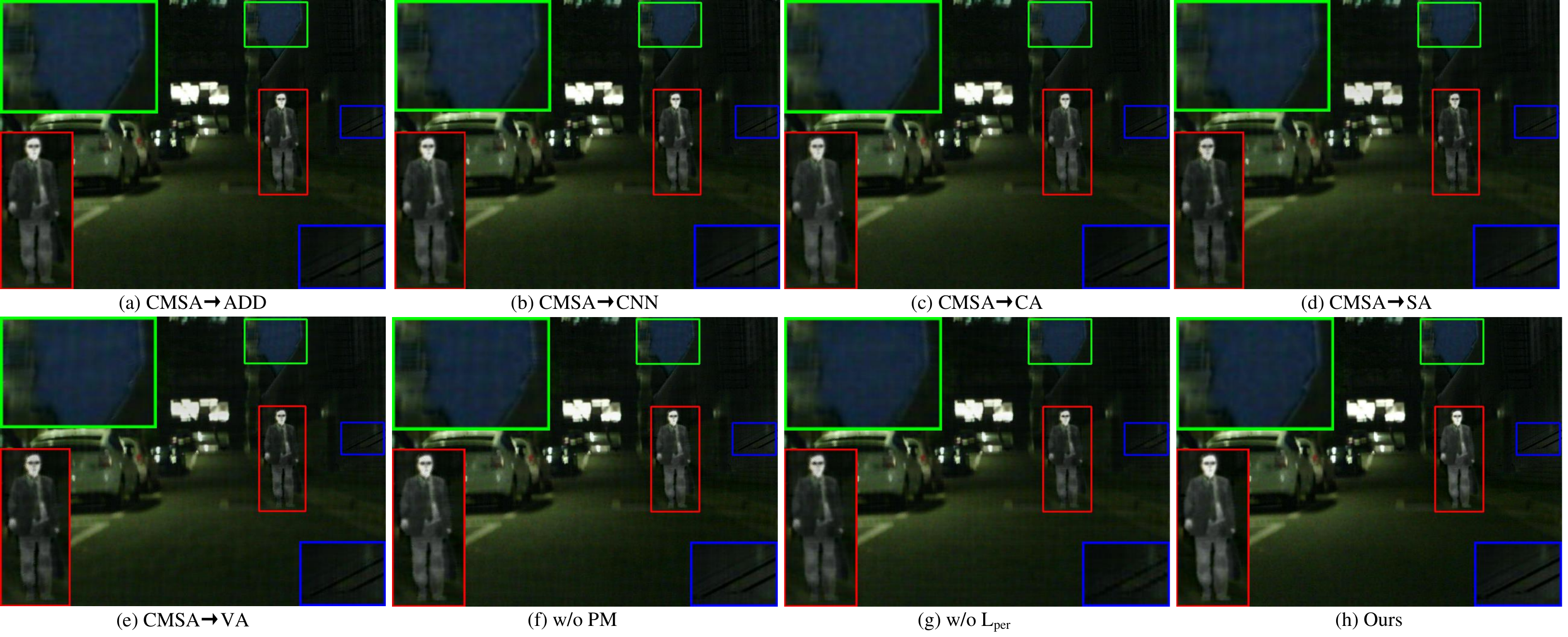}
  \caption{Visualization of the results from each ablation experiment on the image ``00047N" in the MSRS \cite{liu2022target} dataset.}
  \label{fig:ablation}
\end{figure*}

\section{Experiment}
\subsection{Setup}
\subsubsection{Implementation}
Existing datasets suffer from distributional biases, yet salient information is diverse. To address this issue, we amalgamate the training sets of MSRS \cite{tang2022piafusion}, 70\% of M3FD \cite{liu2022target}, and 70\% of RoadScene \cite{xu2020aaai} to form a sufficiently diversified dataset for training, with the remaining portions reserved for testing and validation. All divisions adopt a random splitting method to ensure that the data are uniformly and effectively partitioned. Furthermore, to validate the robust generalization capability of S4Fusion, experiments are conducted on the TNO dataset, which is not part of the training data and contains image styles that differ significantly from those in the training set.

Hyperparameters $\alpha_1, \alpha_2$, and $\alpha_3$ are set to 15, the number of layers $N$ to 3, the number of CMSAB $K$ to 3, and the number of VSS blocks in both encoders and decoder to $G=[1, 2, 1]$. S4Fusion has three layers with channel numbers of [48, 96, 192], a patch size of $4\times4$, and adjacent patches overlapping by one pixel. All experiments are conducted on an RTX 3090 Ti using AdamW as the optimizer and WarmupCosine for learning rate scheduling, with an initial learning rate of 0.0003, peaking at 0.0015, and decaying to 0.0006. The batch size is set to 20, and training is performed for 800 epochs. Input images are randomly cropped to $229\times229$ and augmented with random flipping.

We compare our approach with nine state-of-the-art methods, namely U2Fusion \cite{xu2022u2fusion}, DeFusion \cite{liang2022fusion}, DDFM \cite{zhao2023ddfm}, MambaDFuse \cite{li2024mambadfuse}, SemLA \cite{xie2023semantics}, IRFS \cite{Wang_2023_IF}, CrossFuse \cite{li2024crossfuse}, MAEFuse \cite{li2025maefuse}, and EMMA \cite{zhao2023equivariant}.  Categorized by their interaction module types, IRFS, CrossFuse, MAEFuse, and EMMA are attention-based state-of-the-art (SOTA) fusion methods, while MambaDFuse is a Mamba-based interaction method inspired by the standard cross attention. The remaining methods are all based on CNN interactions. In terms of model fusion scale, all methods except U2Fusion are multi-level fusion approaches.

\subsubsection{Metrics}
For quantitative evaluation, two categories of metrics are employed to assess the effectiveness of S4Fusion. The first category consists of traditional image fusion metrics. Following previous work \cite{li2018densefuse, liu2023sgfusion, zhao2023equivariant}, four metrics are utilized: Spatial Frequency (SF) \cite{eskicioglu1995image}, which measures the rate of change in image grayscale (a larger spatial frequency indicates a clearer image); Average Gradient (AG), which measures the richness of image texture details; Visual Information Fidelity (VIF) \cite{sheikh2006image}, which assesses image information fidelity based on natural scene statistical models; and Qabf \cite{piella2003new}, which reflects the quality of visual information obtained from the input image fusion. Among these, SF and AG are no-reference metrics, while VIF and Qabf are full-reference metrics. The second category comprises perceptual image quality assessment metrics, for which two SOTA metrics are adopted: TOPIQ \cite{chen2024topiq} and MUSIQ \cite{ke2021musiq}. Both leverage pre-trained neural networks to simulate human visual perception. Specifically, they are typically highly sensitive to noise, distortion, and illumination in images. Due to the inherent interpretability challenges of neural networks and the complexity of human preferences, they do not possess a clear preferential explanation like the AG metric. Instead, they comprehensively assess image quality from multiple angles. All bold and red values indicate the best results, while underlined values represent the second-best results.

\subsubsection{Downstream Task}
For the downstream task, we utilize mAP@50 on the M3FD for multi-modal object detection task evaluation. we use YOLOv5 with a pre-trained YOLOv5s model. The fusion results on M3FD are split 7:1:2 for training, validation, and testing, respectively. All methods are evaluated under identical experimental conditions, including the same random seeds, learning rates, and dataset splits. Each method is trained for 40 epochs on its fusion images using YOLOv5, with the best model selected based on validation performance and evaluated on the test set.

\subsection{Qualitative Comparison}
As shown in Figs.\ref{fig:msrs}, \ref{fig:m3fd2}, and \ref{fig:roadscene}, our method effectively highlights salient targets from infrared images. In Fig.\ref{fig:msrs}, we preserve pedestrians obscured by strong light (green box) while retaining ground texture details (red box). In Fig.\ref{fig:m3fd2}, we simultaneously highlight pedestrians and vehicles, preventing them from being obscured by dense fog (red and green boxes). Finally, in Fig.\ref{fig:roadscene}, our method emphasizes a person obscured by fog (green box) and trees with blurred edges (red box). Other methods, such as MambaDFuse, EMMA, SemLA, CrossFuse, and DeFusion, lack the ability to selectively focus on salient information spatially. This leads to target loss in the aforementioned images.

While DDFM, U2Fusion, MAEFuse, and IRFS can better maintain information in these scenarios without being interfered with by strong visible light or fog, they still lose critical information when faced with infrared noise. In Fig.\ref{fig:m3fd1}, they all lose important information on the road sign due to infrared thermal radiation noise (green box) and interfere with already clear salient targets (red box). This is because these methods often rigidly favor information from a specific modality rather than adaptively integrating it. In contrast, even when infrared data is noisy, our method does not absolutely prioritize infrared information but instead preserves truly critical and useful information in the fused image.

Furthermore, our method demonstrates strong generalization capabilities. As shown in Fig.\ref{fig:TNO}, even on datasets outside the training distribution, our approach can still preserve better salient information (green box) and rich texture details (blue box). This is thanks to the excellent simultaneous spatial and modal interaction capabilities of CMSA, which enable our approach to consider spatial information from both modalities during modal interaction. This facilitates a global spatial perspective on information retention and discarding, ultimately achieving the ability to preserve critical information under extreme conditions.

We further quantitatively demonstrate these visually observed effects. In all these cases, sharper edges and clearer images indicate higher image quality. Therefore, we use two no-reference metrics, Average Gradient (AG) and Spatial Frequency (SF), to quantitatively describe our observed results. As shown in Figs.\ref{fig:msrs}, \ref{fig:m3fd1}, \ref{fig:m3fd2}, \ref{fig:roadscene}, and \ref{fig:TNO}, our method consistently exhibits higher SF and AG compared to other methods. It can also be observed that both the SF and AG of our method surpass those of the two original images, which signifies that our approach successfully integrates critical information from each source image while effectively filtering out redundant information.

\begin{table*}[!ht]
  \caption{Quantitative Comparison on the MSRS\cite{tang2022piafusion} dataset.}
  \label{tab:cmp_1}
  \centering
  \tabcolsep=0.68cm
\begin{tabular}{c|c|cccccc}
\hline
Method & year & SF & AG & VIF & Qabf & MUSIQ & TOPIQ \\ \hline
DeFusion & 2022 & 8.606 & 2.781 & 0.763 & 0.530 & 33.033 & 0.214 \\
U2Fusion & 2022 & 6.211 & 1.983 & 0.520 & 0.289 & 32.799 & 0.225 \\
SemLA & 2023 & 6.339 & 2.240 & 0.610 & 0.290 & 28.287 & 0.188 \\
DDFM & 2023 & 7.388 & 2.522 & 0.743 & 0.474 & 32.747 & 0.217 \\
MambaDFuse & 2024 & 11.018 & 3.632 & \textcolor{blue}{\underline{0.985}} & \textcolor{blue}{\underline{0.652}} & 35.958 & 0.231 \\ \hline
IRFS & 2023 & 9.888 & 3.155 & 0.735 & 0.477 & \textcolor{blue}{\underline{36.441}} & 0.224 \\
CrossFuse & 2024 & 9.604 & 3.002 & 0.835 & 0.556 & \textcolor{red}{\textbf{36.843}} & 0.231 \\
EMMA & 2024 & \textcolor{blue}{\underline{11.559}} & \textcolor{blue}{\underline{3.788}} & 0.974 & 0.643 & 35.970 & \textcolor{blue}{\underline{0.232}} \\
MAEFuse & 2025 & 9.569 & 3.461 & 0.747 & 0.502 & 29.275 & 0.227 \\ \hline
S4Fusion & - & \textcolor{red}{\textbf{11.680}} & \textcolor{red}{\textbf{3.793}} & \textcolor{red}{\textbf{1.002}} & \textcolor{red}{\textbf{0.678}} & 35.317 & \textcolor{red}{\textbf{0.247}} \\ \hline
\end{tabular}
\end{table*}

\begin{table*}[!ht]
  \caption{Quantitative Comparison on the RoadScene \cite{xu2020aaai} dataset.}
  \label{tab:cmp_2}
  \centering
  \tabcolsep=0.68cm
\begin{tabular}{c|c|cccccc}
\hline
Method & year & SF & AG & VIF & Qabf & MUSIQ & TOPIQ \\ \hline
DeFusion & 2022 & 8.817 & 3.552 & 0.506 & 0.368 & 40.087 & 0.228 \\
U2Fusion & 2022 & 12.410 & 5.048 & 0.536 & 0.489 & 42.423 & 0.249 \\
SemLA & 2023 & 13.863 & 4.256 & 0.488 & 0.301 & 33.729 & 0.200 \\
DDFM & 2023 & 11.412 & 4.658 & 0.095 & 0.147 & 43.881 & \textcolor{blue}{\underline{0.279}} \\
MambaDFuse & 2024 & 13.623 & 5.279 & \textcolor{blue}{\underline{0.625}} & \textcolor{blue}{\underline{0.508}} & 42.611 & 0.241 \\ \hline
IRFS & 2023 & 10.207 & 3.925 & 0.569 & 0.424 & 41.025 & 0.251 \\
CrossFuse & 2024 & 15.483 & 6.053 & 0.579 & 0.358 & \textcolor{red}{\textbf{45.354}} & \textcolor{red}{\textbf{0.292}} \\
EMMA & 2024 & \textcolor{blue}{\underline{15.816}} & \textcolor{blue}{\underline{6.138}} & 0.624 & 0.443 & 40.658 & 0.236 \\
MAEFuse & 2025 & 12.226 & 5.206 & 0.563 & 0.461 & 38.052 & 0.222 \\ \hline
S4Fusion & - & \textcolor{red}{\textbf{16.156}} & \textcolor{red}{\textbf{6.239}} & \textcolor{red}{\textbf{0.789}} & \textcolor{red}{\textbf{0.611}} & \textcolor{blue}{\underline{44.057}} & 0.263 \\ \hline
\end{tabular}
\end{table*}

\begin{table*}[!ht]
  \caption{Ablation experiments on the TNO Dataset}
  \label{tab:cmp_4}
  \centering
  \tabcolsep=0.68cm
\begin{tabular}{c|c|cccccc}
\hline
Method & year & SF & AG & VIF & Qabf & MUSIQ & TOPIQ \\ \hline
DeFusion & 2022 & 5.652 & 2.298 & 0.509 & 0.332 & 26.830 & 0.213 \\
U2Fusion & 2022 & 8.024 & 3.363 & 0.529 & 0.427 & \textcolor{red}{\textbf{31.554}} & 0.226 \\
SemLA & 2023 & 8.415 & 2.737 & 0.435 & 0.258 & 27.743 & 0.207 \\
DDFM & 2023 & 7.852 & 3.183 & 0.258 & 0.227 & 28.872 & 0.201 \\
MambaDFuse & 2024 & 8.875 & 3.516 & 0.691 & \textcolor{blue}{\underline{0.483}} & 27.891 & 0.214 \\ \hline
IRFS & 2023 & 8.347 & 2.950 & 0.563 & 0.416 & 30.501 & 0.223 \\
CrossFuse & 2024 & 9.973 & 3.71 & \textcolor{blue}{\underline{0.752}} & 0.454 & 30.898 & \textcolor{blue}{\underline{0.251}} \\
EMMA & 2024 & \textcolor{blue}{\underline{10.132}} & \textcolor{blue}{\underline{4.202}} & 0.690 & 0.432 & 28.694 & 0.213 \\
MAEFuse & 2025 & 7.880 & 3.534 & 0.528 & 0.392 & 27.239 & 0.195 \\ \hline
S4Fusion & - & \textcolor{red}{\textbf{11.407}} & \textcolor{red}{\textbf{4.288}} & \textcolor{red}{\textbf{0.803}} & \textcolor{red}{\textbf{0.587}} & \textcolor{blue}{\underline{31.464}} & \textcolor{red}{\textbf{0.302}} \\ \hline
\end{tabular}
\end{table*}

\begin{table*}[!ht]
  \caption{Quantitative Comparison on the M3FD\cite{liu2022target} Dataset}
  \label{tab:cmp_3}
  \centering
  \tabcolsep=0.68cm
\begin{tabular}{c|c|cccccc}
\hline
Method & year & SF & AG & VIF & Qabf & MUSIQ & TOPIQ \\ \hline
DeFusion & 2022 & 7.752 & 2.789 & 0.529 & 0.337 & 51.123 & 0.325 \\
U2Fusion & 2022 & 11.211 & 4.259 & 0.606 & 0.532 & 50.105 & 0.331 \\
SemLA & 2023 & 7.370 & 2.748 & 0.377 & 0.179 & 35.542 & 0.251 \\
DDFM & 2023 & 9.583 & 3.381 & 0.491 & 0.384 & 50.862 & 0.336 \\
MambaDFuse & 2024 & 12.445 & 4.502 & 0.668 & 0.534 & 51.330 & 0.334 \\ \hline
IRFS & 2023 & 10.920 & 3.684 & 0.621 & 0.503 & 52.439 & 0.357 \\
CrossFuse & 2024 & 12.210 & 4.181 & 0.740 & 0.575 & \textcolor{red}{\textbf{54.689}} & 0.376 \\
EMMA & 2024 & \textcolor{red}{\textbf{15.882}} & \textcolor{red}{\textbf{5.719}} & \textcolor{blue}{\underline{0.751}} & \textcolor{blue}{\underline{0.592}} & 52.522 & 0.345 \\
MAEFuse & 2025 & 9.881 & 3.994 & 0.515 & 0.382 & 42.852 & 0.287 \\ \hline
S4Fusion & - & \textcolor{blue}{\underline{15.363}} & \textcolor{blue}{\underline{5.312}} & \textcolor{red}{\textbf{0.910}} & \textcolor{red}{\textbf{0.661}} & \textcolor{blue}{\underline{53.783}} & \textcolor{red}{\textbf{0.377}} \\ \hline
\end{tabular}
\end{table*}

\begin{table*}[!ht]
  \caption{Ablation experiments on the MSRS\cite{liu2022target} Dataset}
  \label{tab:abl}
  \centering
  \tabcolsep=0.65cm
\begin{tabular}{cc|cccccc}
\hline
Number & Method & AG & VIF & Qabf & SF & MUSIQ & TOPIQ \\ \hline
Exp.1 & CMSA$\to$ ADD & 3.782 & 0.843 & 0.614 & 11.52 & 35.103 & 0.231 \\
Exp.2 & CMSA$\to$ CNN & 3.730 & 0.919 & 0.631 & 11.669 & 34.973 & 0.242 \\
Exp.3 & CMSA$\to$ CA & 3.621 & 0.998 & 0.679 & 11.679 & 35.212 & 0.244 \\
Exp.4 & CMSA$\to$ SA & 3.539 & 0.663 & 0.472 & 11.622 & 34.918 & 0.233 \\
Exp.5 & CMSA$\to$ VA & 3.744 & 0.784 & 0.671 & 11.471 & 34.662 & 0.240 \\
Exp.6 & w/o PM & 3.783 & 0.851 & 0.637 & 11.493 & 35.211 & 0.232 \\
Exp.7 & w/o \(\mathcal{L}_{\mathrm{per}}\) & 3.776 & 0.970 & 0.652 & 11.036 & 35.012 & 0.241 \\ \hline
Exp.8 & Ours & \textcolor{red}{\textbf{3.797}} & \textcolor{red}{\textbf{1.002}} & \textcolor{red}{\textbf{0.681}} & \textcolor{red}{\textbf{11.685}} & \textcolor{red}{\textbf{35.317}} & \textcolor{red}{\textbf{0.247}} \\ \hline
\end{tabular}
\end{table*}

\begin{table}[!ht]
  \caption{mAP@50 values for MM Detection on M3FD\cite{liu2022target} Dataset. All the methods being compared are abbreviated using the first three letters to save space.}
  \label{tab:dete}
  \tabcolsep=0.05cm
  \centering
\begin{tabular}{c|ccccccc|cccc|c}
\hline
 & IR & VI & DeF & DDF & Sem & U2F & Mam & IRF & Cro & EMM & MAE & S4F \\
mAP50 & 73.8 & 79.3 & 80.5 & 80.4 & 77.2 & 80.1 & 79.6 & \textcolor{blue}{\underline{81.0}} & 80.2 & 80.4 & 80.0 & \textcolor{red}{\textbf{81.3}} \\ \hline
\end{tabular}
\end{table}

 
\subsection{Quantitative Comparison}
\label{sec:qua}
As shown in Table.\ref{tab:cmp_1}, \ref{tab:cmp_2}, \ref{tab:cmp_3}, and \ref{tab:cmp_4}, we conduct quantitative testing on four datasets. Overall, compared to other methods, our method yield outstanding results. S4Fusion demonstrates higher AG and SF, suggesting superior preservation of texture and structural information. Additionally, it also achieves good results in VIF and Qabf, indicating that it preserves more original image information while simultaneously capturing more salient information. S4Fusion achieves competitive results on MUSIQ and TOPIQ, which means it boasts better visual effects.

Specifically, our method performs best on the RoadScene, TNO, and MSRS datasets, achieving leading positions across almost all metrics. In contrast, S4Fusion leading position is slightly diminished on the M3FD dataset. This is highly likely due to the inherent structural limitations of the SSSM itself. The images in the M3FD dataset possess exceptionally high resolution. Since SSSM primarily captures context through hidden states, and the information storage capacity of these hidden states is inherently limited, SSSM's ability to model global information tends to decrease when resolutions become excessively large. Directly increasing the dimension of the hidden states, however, could lead to increased retention of redundant information and higher computational costs. Therefore, a dynamically variable capacity for the hidden layer might be a promising area for in-depth research in the future.

\subsection{multi-modal Object Detection}
A multi-modal object detection task was conducted to verify whether our method can effectively highlight salient targets. Qualitatively, as shown in Fig.\ref{fig:detect}, our method did not lose salient targets. Other methods, such as EMMA, DDFM, CrossFuse, and DeFusion, were less effective at highlighting small targets, which led to the detector failing to detect key objects (the traffic light in the green box and the person in the cyan box). Furthermore, compared to methods like MambaDFuse, IRFS, and SemLA, which are relatively effective at highlighting small targets, our detection results had higher confidence scores, meaning our method's saliency is more pronounced and better suited for downstream tasks.

Quantitatively, as shown in Table.\ref{tab:dete}, S4Fusion achieves a higher mAP@50, signifying superior average detection accuracy. Other approaches, such as IRFS, EMMA, DeFusion, and DDFM, also exhibit relatively high accuracy, which is consistent with the qualitative visual comparison results. However, methods like MAEFuse, MambaDFuse, and SemLA show poorer results, partly due to insufficiently prominent saliency, which aligns with the quantitative description in the qualitative experiments.

\begin{table}[!ht]
  \caption{Comparison of FLOPs and parameters for various methods on the MSRS test set. All the methods being compared are abbreviated using the first three letters to save space.}
  \label{tab:params}
  \tabcolsep=0.025cm
  \centering
\begin{tabular}{c|ccccc|cccc|c}
\hline
 & DEF & DDF & Sem & U2F & Mam & IRF & Cro & EMM & MAE & S4F \\ \hline
GFLOPs$\downarrow$ & 80.13 & 664k & 66.45 & 52.30 & 55.09 & 73.73 & 134.85 & \textcolor{red}{\textbf{41.53}} & 1004.50 & \textcolor{blue}{\underline{48.93}} \\
Params(M)$\downarrow$ & 7.87 & 552.70 & 7.27 & \textcolor{blue}{\underline{0.86}} & 5.32 & \textcolor{red}{\textbf{0.24}} & 20.33 & 1.51 & 325.02 & 2.030 \\ \hline
\end{tabular}
\end{table}

\subsection{Efficiency Comparison}
To verify whether S4Fusion could leverage the linear complexity advantage of SSSM, a qualitative comparison of model efficiency is conducted. As shown in Table.\ref{tab:params}, S4Fusion exhibits lower parameters and computational complexity. Compared to Transformer-based methods like MAEFuse, IRFS, and CrossFuse, it demonstrates significantly lower computational requirements, proving its superiority in processing speed. However, S4Fusion's computational cost is slightly higher than that of EMMA, which also uses a Transformer, primarily because EMMA employs a lightweight attention mechanism, leading to higher inherent efficiency. In comparison to the similar Mamba-based method, MambaDFuse, S4Fusion has a clear advantage with fewer parameters and lower computational demands. Overall, S4Fusion achieves relatively low computational complexity and parameter count, leveraging the linear complexity advantage of SSSM.

\subsection{Ablation Study}
For the module design, in Exp.1-5, we successively replace the proposed CMSA with five different modules: ADD, CNN, standard channel cross-attention (CA) \cite{zamir2022restormer}, standard spatial cross-attention (SA) \cite{li2024crossfuse}, and vanilla attention (VA) \cite{zhao2023equivariant}. For the CNN and standard visual attention modules, infrared and visible features are concatenated along the channel dimension before input. We control the number of layers in these modules to maintain a similar parameter count to CMSA after replacement. Among these, ADD directly sums the two features, lacking inter-modality interaction. As shown in CMSA → ADD in Fig.\ref{fig:ablation}, this results in significant information loss and poor image quality. As depicted in CMSA → CNN in Fig.\ref{fig:ablation}, CNN suffers from a limited global receptive field, leading to the loss of critical information in the fusion results.

As shown in CMSA → CA in Fig.\ref{fig:ablation}, standard channel cross-attention achieves results closest to S4Fusion, but its lack of selective focus on local regions leads to deficiencies in highlighting salient objects. In CMSA → SA in Fig.\ref{fig:ablation}, standard spatial cross-attention yields poorer results, with a noticeable reduction in the prominence of human figures. This might be due to its less effective adaptation to low-level tasks compared to CA, and an inconsistency with the distribution of features extracted by SSSM. However, even in qualitative comparisons, the fusion results of CrossFuse, which uses standard spatial cross-attention and matches the backbone distribution, are still inferior to S4Fusion. Finally, in CMSA → VA in Fig.\ref{fig:ablation}, since the two modalities are merely concatenated as input and interacted only via a self-attention mechanism on features, it struggles to focus on matching information from the other modality, ultimately leading to fusion results that tend to converge towards the average of the two modalities.

In Exp.6, the Patch Mark is removed to assess its role in distinguishing spatial and inter-modality interactions within CMSA. The results indicate that without the Patch Mark, CMSA becomes unable to differentiate between inter-modal and spatial interactions during scanning, leading to spatial distortions in the fused image.

In Exp.7, the proposed \(\mathcal{L}_{\mathrm{per}}\) function is removed while maintaining the remaining experimental configurations to verify its effectiveness. The results confirm the efficacy of this perceptual loss. As shown in w/o \(\mathcal{L}_{\mathrm{per}}\) in Fig.\ref{fig:ablation}, when the loss function is omitted, the amount of salient information in the fused image is significantly reduced, resulting in compromised image quality.

\begin{figure}[!tb]
  \centering
   \includegraphics[width=0.45\textwidth,height=0.25\textwidth]{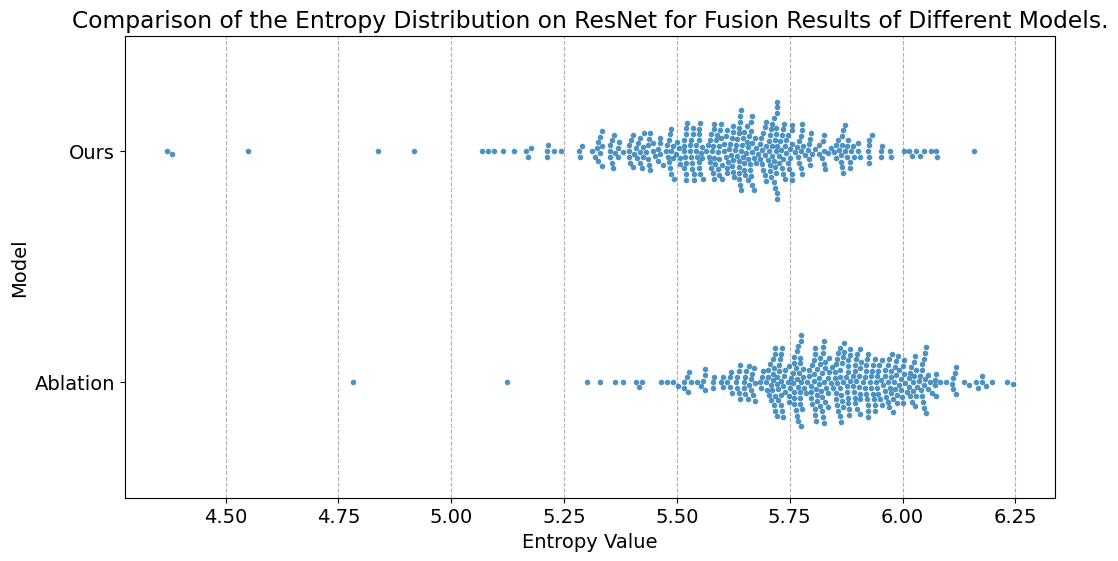}
  \caption{The entropy distribution of our model's and the w/o $\mathcal{L}_{\mathrm{per}}$ version's fusion results on the MSRS dataset, at the pre-trained ResNet's decision layer. It reveals that without our proposed enhancement loss, the entropy distribution is generally higher. This indicates a flatter ResNet decision layer distribution and less prominent salient targets.}
  \label{fig:abl_en}
\end{figure}

\subsubsection{Further Analysis of \(\mathcal{L}_{\mathrm{per}}\)}
To further analyze the principle of $\mathcal{L}_{\mathrm{per}}$,  more in-depth theoretical analysis and experiments are conducted. 

First, theoretically, the decision layer of a pre-trained network contains perceptual information regarding salient targets. As discovered by Arandjelovic \textit{et al.} \cite{arandjelovic2017look}, each of the 512 channels after the last pooling layer of a convolutional network represents category-specific information, and visualizations of their feature maps also demonstrate that each channel corresponds to a particular object of interest. 

Similarly, this phenomenon has also been observed in Transformer-based models. In DINO \cite{caron2021emerging}, the attention map visualizations of different heads of the [CLS] token in the last layer's attention correspond to the semantic regions of different objects in the image. These observations indicate that the decision layer is capable of identifying salient regions within an image. 

For classic classification networks like ResNet, the output of their decision layer is transformed into a probability distribution via softmax. Naturally, by manipulating this probability distribution, the salient regions in the image can be enhanced. To achieve this, we should increase the probability values corresponding to existing salient targets in the image. This implies that the probability distribution should become sharper, \textit{i.e.}, possess smaller information entropy. This explains the mechanism of action for $\mathcal{L}_{\mathrm{per}}$.

However, whether this loss function truly converges as intended remains a question. To verify this, we visualized the information entropy distribution of our method and the w/o $\mathcal{L}_{\mathrm{per}}$ model at the ResNet decision layer on the MSRS dataset. As shown in Fig.\ref{fig:abl_en}, the information entropy distribution of our method is generally shifted more to the left, meaning our distribution is sharper, and salient regions are more prominent. For the ablation model not using $\mathcal{L}_{\mathrm{per}}$, due to the absence of the perceptual enhancement loss constraint, the distribution of critical information in its fused image is more dispersed, and saliency is weaker.
\begin{figure*}[!tb]
  \centering
   \includegraphics[width=1\textwidth,height=0.16\textwidth]{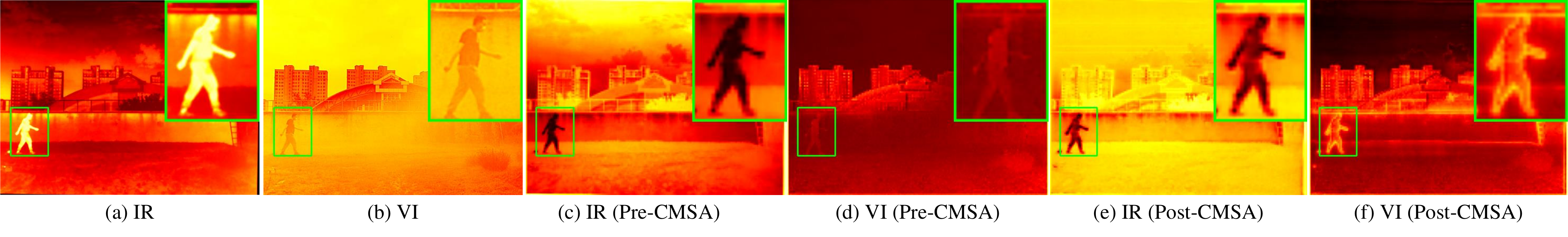}
  \caption{Visualization of the feature maps before and after the second-layer CMSA in M3FD \cite{liu2022target} for "00388", along with the original images.}
  \label{fig:visual}
\end{figure*}

\subsection{Visualization}
To investigate how CMSA operates, the features pre- and post-processing are analyzed. Fig.\ref{fig:visual} illustrates the changes observed. Prior to CMSA processing, visible features displayed a blurry human figure and an unclear background, whereas infrared features showed the opposite characteristics. Following CMSA processing, infrared features highlighted the human figure against the background, while both human figures and backgrounds in the visible image were improved. This showcases the CMSA module's capacity to capture and integrate critical information from different modalities.

\subsection{Interpretation}
\label{Interpretation}
In this section, the parameters of Cross SS2D are further explained in detail.  First, we elaborate on the operation of \(\mathrm{Inter}\) in Cross SS2D. Taking $\mathrm{Inter}(S_{\mathrm{B_mI}}, S_{\mathrm{B_mV}})(\textbf{x})$ as an example, the input sequence is assumed to be $\textbf{x} = [I_{1}, V_{1}, I_{2}, V_{2}, ..., I_{L}, V_{L}]$, where $V_{i}$ and $I_{i}$ represent the $i$-th patches of the visible and infrared modalities, respectively, for $i \in \{1, 2, ..., L\}$.

The \(\mathrm{Inter}\) operation arranges the parameters into a sequence of length 2L: \(\mathrm{Inter}(S_{\mathrm{B_mI}}, S_{\mathrm{B_mV}}) = [S_{\mathrm{B_mI}}, S_{\mathrm{B_mV}}, S_{\mathrm{B_mI}}, S_{\mathrm{B_mV}}, ..., S_{\mathrm{B_mI}}, S_{\mathrm{B_mV}}]\), where the $S_{B_mI}$ and $S_{B_mV}$ are linear layers. For the linear layers, the \(\mathrm{Inter}\) operation simply aligns the same modality's linear layer with the sequence positions: 
\begin{equation}
\begin{aligned}
\mathrm{Inter}(S_{\mathrm{B_mI}}, S_{\mathrm{B_mV}})(\textbf{x}) = [S_{\mathrm{B_mI}}(I_{1}), S_{\mathrm{B_mV}}(V_{1}), \\ S_{\mathrm{B_mI}}(I_{2}), S_{\mathrm{B_mV}}(V_{2}), ..., S_{\mathrm{B_mI}}(I_{L}), S_{\mathrm{B_mV}}(V_{L})]
\end{aligned}
\end{equation}
It means that the corresponding patches are input into their respective linear layers to compute the corresponding $B_m$ parameters.

\subsubsection{Interpretation of \(A_m\)}
According to the theory of parameter \(A_m\) in Mamba, parameter \(A_m\) can also become selective. However, following Mamba's design philosophy, we make parameter \(A_m\) independent of the input and shared between the two modalities. Based on the discretization formula $\bar{A}_m = \exp(\Delta A_m)$, making $\Delta$ selective is sufficient for $\bar{A}_m$ to be selective, thus making \(A_m\) selective is unnecessary. Similarly, for two different modalities, it is not required to set separate parameter \(A_m\) for each; the respective parameters $\Delta$ are sufficient to distinguish their respective $\bar{A}_m$ for different modalities.
\begin{table}[tb]
\centering
\caption{Comparison of the average $\Delta$ for the two modalities across different layers.}
\label{delta}
\begin{tabular}{c|cc}
       & \(\bar{\Delta}_{\mathrm{I}}\)  & \(\bar{\Delta}_{\mathrm{V}}\)   \\ \hline
Layer1 & 0.0014 & -0.0056 \\
Layer2 & -3.999 & -5.885  \\
Layer3 & -6.308 & -19.811 \\ \hline
\end{tabular}
\end{table}
\subsubsection{Interpretation of $\Delta$}
Similar to parameter $\textbf{\(B_m\)}$, $\Delta$ can be expressed as $\Delta = [\Delta_{\mathrm{I}1}, \Delta_{\mathrm{V}1}, \Delta_{\mathrm{I}2}, \Delta_{\mathrm{V}2}, ..., \Delta_{\mathrm{I}L}, \Delta_{\mathrm{V}L}]$. Generally, $\Delta$ controls the relationship between the current and previous inputs. In SSM, $\Delta$ represents the interval after discretizing a continuous signal. Larger values of $\Delta$ indicate a greater emphasis on the current input, while smaller values indicate the opposite. Notably, the infrared parameter $\Delta_{\mathrm{I}}$ controls the interval for different spatial positions of different modalities, whereas $\Delta_{\mathrm{V}}$ controls the interval for the same spatial position of different modalities. 

We denote $\bar{\Delta}_{\mathrm{I}}$ and $\bar{\Delta}_{\mathrm{V}}$ as the averages of the $\Delta$ parameters for infrared and visible from our trained model. Let $\Delta_{\mathrm{V}j}^k$ be the distance between the $j$-th visible modality element and the previous infrared modality element in the $k$-th CMSAB of a certain layer, and similarly for $\Delta_{\mathrm{I}j}^k$. Then $\bar{\Delta}_{\mathrm{I}}$ and $\bar{\Delta}_{\mathrm{V}}$ can be formally expressed as:
\begin{gather}
\bar \Delta_{I}=\frac{1}{LK}\sum\limits_{k=1}^K\sum\limits_{j=1}^L\Delta_{\mathrm{I}j}^k, \bar \Delta_{V}=\frac{1}{LK}\sum\limits_{k=1}^K\sum\limits_{j=1}^L\Delta_{\mathrm{V}j}^k
\end{gather}
As shown in Table.\ref{delta}, we conducted experiments on the MSRS test set, calculating the average $\Delta$ parameter values for infrared and visible across all CMSABs in each layer to obtain $\bar{\Delta}_{\mathrm{I}}$ and $\bar{\Delta}_{\mathrm{V}}$. Due to the frequent inter-modal and spatial interactions in Cross SS2D, this interleaving $\Delta$ effectively controls the transmission of information between modalities and spatial locations.

Most of the time, $\bar{\Delta}_{\mathrm{I}}$ represents the interval between adjacent patches of different modalities in space. As shown in Table.\ref{delta}, in all layers, $\bar{\Delta}_{\mathrm{I}}$ is greater than $\bar{\Delta}_{\mathrm{V}}$. Larger values indicate greater differences between inputs over time, emphasizing current position information and compressing past information to capture spatial information. This is because the information difference between spatially adjacent patches is greater than the patches of different modalities in the same position. For example, infrared and visible images at the same spatial position may capture the same object, exhibiting strong complementarity.

Conversely, $\bar{\Delta}_{\mathrm{V}}$ represents the interval between patches of the same position but different modalities. These values are significantly smaller than $\bar{\Delta}_{\mathrm{I}}$, indicating that the model prefers previous information and selects a smaller portion of complementary information from the current position. SSMs reconstruct signals better when they are closer, suggesting that $\bar{\Delta}_{\mathrm{V}}$ controls the acquisition of complementary information between patches of different modalities at the same position.

We also observe that the value of both \(\bar{\Delta}_{\mathrm{I}}\) and \(\bar{\Delta}_{\mathrm{V}}\) decreases with increasing layers. This suggests that, as the network layers deepen, the emphasis shifts more towards the current state because deeper features typically carry richer semantic information. In deep feature maps, only semantic information is usually highlighted, while semantically irrelevant information is discarded. This leads to a significant portion of the image having lower activation values, and consequently, the mean of $\Delta$ tends to decrease. This phenomenon suggests that in deeper networks, modal interaction focuses more on semantic information, and most non-semantic information is discarded, which benefits the prominence of critical information.

\subsubsection{Interpretation of \(B_m\) and \(C_m\)}
\(B_m\) and \(C_m\) regulate the input-to-hidden state and hidden state-to-output information flow, respectively. We employ distinct parameters for each modality to compute \(B_m\) and \(C_m\). This facilitates the precise extraction and retention of complementary modality and spatial information from the hidden state, tailored to the specific characteristics of each modality's information.

\section{Conclusion}
This paper introduces S4Fusion, a novel approach designed to adaptively retain significant information without relying on manually annotated labels. To achieve this, we first propose a one-stage fusion module, CMSA, which enhances the integration of complementary information across modalities at a global level. Unlike previous methods, CMSA effectively addresses the deficiency of global spatial information during inter-modal interaction. Furthermore, to enable our model to perceive salient objects, we design a novel loss function, $\mathcal{L}_{\mathrm{per}}$ This function leverages a pre-trained model to minimize uncertainty in fused images, thereby facilitating the adaptive retention of salient information. Extensive experiments validate the effectiveness of our proposed approach.

Despite the excellent results achieved by our method, some limitations still exist. First, as mentioned in Sec.\ref{sec:qua}, due to the inherent nature of SSSM, its global modeling capability is determined by the size of its hidden layer. In image fusion, input images vary widely in resolution, and whether SSSMs can dynamically adjust its hidden layer capacity for different resolutions is an interesting future research direction.

Additionally, there is still room for further improvement in our method's speed and parameter count. For instance, the Linear module within SSSM might be replaced by a more lightweight $1 \times 1$ convolution. Exploring these potentially suitable lightweight structures for SSSMs and fusion tasks is also worth investigating in the future.

Finally, regarding the $\mathcal{L}_{\mathrm{per}}$, we empirically selected ResNet50 as the pre-trained model. We acknowledge, however, that this pre-trained model inherently possesses certain biases in its feature distribution. Therefore, we believe that exploring a broader range of pre-trained models with diverse architectures, and investigating their potential integration for a more robust and unbiased perceptual loss calculation, presents a promising avenue for future research.


\bibliographystyle{IEEEtran}
\bibliography{tip}

\begin{thebibliography}{10}
\providecommand{\url}[1]{#1}
\csname url@samestyle\endcsname
\providecommand{\newblock}{\relax}
\providecommand{\bibinfo}[2]{#2}
\providecommand{\BIBentrySTDinterwordspacing}{\spaceskip=0pt\relax}
\providecommand{\BIBentryALTinterwordstretchfactor}{4}
\providecommand{\BIBentryALTinterwordspacing}{\spaceskip=\fontdimen2\font plus
\BIBentryALTinterwordstretchfactor\fontdimen3\font minus \fontdimen4\font\relax}
\providecommand{\BIBforeignlanguage}[2]{{%
\expandafter\ifx\csname l@#1\endcsname\relax
\typeout{** WARNING: IEEEtran.bst: No hyphenation pattern has been}%
\typeout{** loaded for the language `#1'. Using the pattern for}%
\typeout{** the default language instead.}%
\else
\language=\csname l@#1\endcsname
\fi
#2}}
\providecommand{\BIBdecl}{\relax}
\BIBdecl

\bibitem{liu2023sgfusion}
J.~Liu, R.~Dian, S.~Li, and H.~Liu, ``Sgfusion: A saliency guided deep-learning framework for pixel-level image fusion,'' \emph{Information Fusion}, vol.~91, pp. 205--214, 2023.

\bibitem{zhao2020bayesian}
Z.~Zhao, S.~Xu, C.~Zhang, J.~Liu, and J.~Zhang, ``Bayesian fusion for infrared and visible images,'' \emph{Signal Processing}, vol. 177, p. 107734, 2020.

\bibitem{ZhaoDIDFuse2020}
Z.~Zhao, S.~Xu, C.~Zhang, J.~Liu, J.~Zhang, and P.~Li, ``Didfuse: Deep image decomposition for infrared and visible image fusion,'' in \emph{{IJCAI}}.\hskip 1em plus 0.5em minus 0.4em\relax ijcai.org, 2020, pp. 970--976.

\bibitem{zhang2023visible}
X.~Zhang and Y.~Demiris, ``Visible and infrared image fusion using deep learning,'' \emph{IEEE Transactions on Pattern Analysis and Machine Intelligence}, 2023.

\bibitem{zhang2021image}
H.~Zhang, H.~Xu, X.~Tian, J.~Jiang, and J.~Ma, ``Image fusion meets deep learning: A survey and perspective,'' \emph{Information Fusion}, vol.~76, pp. 323--336, 2021.

\bibitem{falahkheirkhah2022drb}
K.~Falahkheirkhah, K.~Yeh, M.~P. Confer, and R.~Bhargava, ``Drb-net: Dilated residual block network for infrared image restoration,'' in \emph{International Symposium on Visual Computing}.\hskip 1em plus 0.5em minus 0.4em\relax Springer, 2022, pp. 104--115.

\bibitem{li2021embedded}
J.~Li, Y.~Peng, and T.~Jiang, ``Embedded real-time infrared and visible image fusion for uav surveillance,'' \emph{Journal of Real-Time Image Processing}, vol.~18, no.~6, pp. 2331--2345, 2021.

\bibitem{sun2021fusion}
H.~Sun, Q.~Liu, J.~Wang, J.~Ren, Y.~Wu, H.~Zhao, and H.~Li, ``Fusion of infrared and visible images for remote detection of low-altitude slow-speed small targets,'' \emph{IEEE Journal of Selected Topics in Applied Earth Observations and Remote Sensing}, vol.~14, pp. 2971--2983, 2021.

\bibitem{li2023feature}
H.~Li, J.~Zhao, J.~Li, Z.~Yu, and G.~Lu, ``Feature dynamic alignment and refinement for infrared--visible image fusion: Translation robust fusion,'' \emph{Information Fusion}, vol.~95, pp. 26--41, 2023.

\bibitem{TANG2023PSFusion}
L.~Tang, H.~Zhang, H.~Xu, and J.~Ma, ``Rethinking the necessity of image fusion in high-level vision tasks: A practical infrared and visible image fusion network based on progressive semantic injection and scene fidelity,'' \emph{Information Fusion}, vol.~99, p. 101870, 2023.

\bibitem{zhao2023metafusion}
W.~Zhao, S.~Xie, F.~Zhao, Y.~He, and H.~Lu, ``Metafusion: Infrared and visible image fusion via meta-feature embedding from object detection,'' in \emph{Proceedings of the IEEE/CVF Conference on Computer Vision and Pattern Recognition}, 2023, pp. 13\,955--13\,965.

\bibitem{wang2023interactively}
D.~Wang, J.~Liu, R.~Liu, and X.~Fan, ``An interactively reinforced paradigm for joint infrared-visible image fusion and saliency object detection,'' \emph{Information Fusion}, vol.~98, p. 101828, 2023.

\bibitem{peng2023siamese}
J.~Peng, H.~Zhao, Z.~Hu, Y.~Zhuang, and B.~Wang, ``Siamese infrared and visible light fusion network for rgb-t tracking,'' \emph{International Journal of Machine Learning and Cybernetics}, vol.~14, no.~9, pp. 3281--3293, 2023.

\bibitem{gu2021efficiently}
A.~Gu, K.~Goel, and C.~R{\'e}, ``Efficiently modeling long sequences with structured state spaces,'' \emph{arXiv preprint arXiv:2111.00396}, 2021.

\bibitem{gu2021combining}
A.~Gu, I.~Johnson, K.~Goel, K.~Saab, T.~Dao, A.~Rudra, and C.~R{\'e}, ``Combining recurrent, convolutional, and continuous-time models with linear state space layers,'' \emph{Advances in neural information processing systems}, vol.~34, pp. 572--585, 2021.

\bibitem{Tang_2023_DATFuse}
W.~Tang, F.~He, Y.~Liu, Y.~Duan, and T.~Si, ``Datfuse: Infrared and visible image fusion via dual attention transformer,'' \emph{IEEE Transactions on Circuits and Systems for Video Technology}, vol.~33, no.~7, pp. 3159--3172, 2023.

\bibitem{liang2022fusion}
P.~Liang, J.~Jiang, X.~Liu, and J.~Ma, ``Fusion from decomposition: A self-supervised decomposition approach for image fusion,'' in \emph{European Conference on Computer Vision}.\hskip 1em plus 0.5em minus 0.4em\relax Springer, 2022, pp. 719--735.

\bibitem{li2024mambadfuse}
Z.~Li, H.~Pan, K.~Zhang, Y.~Wang, and F.~Yu, ``Mambadfuse: A mamba-based dual-phase model for multi-modality image fusion,'' \emph{arXiv preprint arXiv:2404.08406}, 2024.

\bibitem{he2016deep}
K.~He, X.~Zhang, S.~Ren, and J.~Sun, ``Deep residual learning for image recognition,'' in \emph{Proceedings of the IEEE conference on computer vision and pattern recognition}, 2016, pp. 770--778.

\bibitem{li2018densefuse}
H.~Li and X.-J. Wu, ``Densefuse: A fusion approach to infrared and visible images,'' \emph{IEEE Transactions on Image Processing}, vol.~28, no.~5, pp. 2614--2623, 2018.

\bibitem{li2020nestfuse}
H.~Li, X.-J. Wu, and T.~Durrani, ``Nestfuse: An infrared and visible image fusion architecture based on nest connection and spatial/channel attention models,'' \emph{IEEE Transactions on Instrumentation and Measurement}, vol.~69, no.~12, pp. 9645--9656, 2020.

\bibitem{li2021rfn}
H.~Li, X.-J. Wu, and J.~Kittler, ``Rfn-nest: An end-to-end residual fusion network for infrared and visible images,'' \emph{Information Fusion}, vol.~73, pp. 72--86, 2021.

\bibitem{ma2019fusiongan}
J.~Ma, W.~Yu, P.~Liang, C.~Li, and J.~Jiang, ``Fusiongan: A generative adversarial network for infrared and visible image fusion,'' \emph{Information fusion}, vol.~48, pp. 11--26, 2019.

\bibitem{ma2020ganmcc}
J.~Ma, H.~Zhang, Z.~Shao, P.~Liang, and H.~Xu, ``Ganmcc: A generative adversarial network with multiclassification constraints for infrared and visible image fusion,'' \emph{IEEE Transactions on Instrumentation and Measurement}, vol.~70, pp. 1--14, 2020.

\bibitem{zhang2020rethinking}
H.~Zhang, H.~Xu, Y.~Xiao, X.~Guo, and J.~Ma, ``Rethinking the image fusion: A fast unified image fusion network based on proportional maintenance of gradient and intensity,'' in \emph{Proceedings of the AAAI conference on artificial intelligence}, vol.~34, no.~07, 2020, pp. 12\,797--12\,804.

\bibitem{zhao2023cddfuse}
Z.~Zhao, H.~Bai, J.~Zhang, Y.~Zhang, S.~Xu, Z.~Lin, R.~Timofte, and L.~Van~Gool, ``Cddfuse: Correlation-driven dual-branch feature decomposition for multi-modality image fusion,'' in \emph{Proceedings of the IEEE/CVF conference on computer vision and pattern recognition}, 2023, pp. 5906--5916.

\bibitem{tang2022image}
L.~Tang, J.~Yuan, and J.~Ma, ``Image fusion in the loop of high-level vision tasks: A semantic-aware real-time infrared and visible image fusion network,'' \emph{Information Fusion}, vol.~82, pp. 28--42, 2022.

\bibitem{xu2022u2fusion}
H.~Xu, J.~Ma, J.~Jiang, X.~Guo, and H.~Ling, ``U2fusion: A unified unsupervised image fusion network,'' \emph{IEEE Transactions on Pattern Analysis and Machine Intelligence}, 2022.

\bibitem{wang2024infrared}
Y.~Wang, L.~Miao, Z.~Zhou, L.~Zhang, and Y.~Qiao, ``Infrared and visible image fusion with language-driven loss in clip embedding space,'' \emph{arXiv preprint arXiv:2402.16267}, 2024.

\bibitem{radford2021learning}
A.~Radford, J.~W. Kim, C.~Hallacy, A.~Ramesh, G.~Goh, S.~Agarwal, G.~Sastry, A.~Askell, P.~Mishkin, J.~Clark \emph{et~al.}, ``Learning transferable visual models from natural language supervision,'' in \emph{International conference on machine learning}.\hskip 1em plus 0.5em minus 0.4em\relax PMLR, 2021, pp. 8748--8763.

\bibitem{arandjelovic2017look}
R.~Arandjelovic and A.~Zisserman, ``Look, listen and learn,'' in \emph{Proceedings of the IEEE international conference on computer vision}, 2017, pp. 609--617.

\bibitem{caron2021emerging}
M.~Caron, H.~Touvron, I.~Misra, H.~J{\'e}gou, J.~Mairal, P.~Bojanowski, and A.~Joulin, ``Emerging properties in self-supervised vision transformers,'' in \emph{Proceedings of the IEEE/CVF international conference on computer vision}, 2021, pp. 9650--9660.

\bibitem{zhao2023equivariant}
Z.~Zhao, H.~Bai, J.~Zhang, Y.~Zhang, K.~Zhang, S.~Xu, D.~Chen, R.~Timofte, and L.~Van~Gool, ``Equivariant multi-modality image fusion,'' \emph{CVPR}, 2024.

\bibitem{liu2022target}
J.~Liu, X.~Fan, Z.~Huang, G.~Wu, R.~Liu, W.~Zhong, and Z.~Luo, ``Target-aware dual adversarial learning and a multi-scenario multi-modality benchmark to fuse infrared and visible for object detection,'' in \emph{Proceedings of the IEEE/CVF Conference on Computer Vision and Pattern Recognition}, 2022, pp. 5802--5811.

\bibitem{li2025maefuse}
J.~Li, J.~Jiang, P.~Liang, J.~Ma, and L.~Nie, ``Maefuse: Transferring omni features with pretrained masked autoencoders for infrared and visible image fusion via guided training,'' \emph{IEEE Transactions on Image Processing}, 2025.

\bibitem{ding2022scaling}
X.~Ding, X.~Zhang, J.~Han, and G.~Ding, ``Scaling up your kernels to 31x31: Revisiting large kernel design in cnns,'' in \emph{Proceedings of the IEEE/CVF conference on computer vision and pattern recognition}, 2022, pp. 11\,963--11\,975.

\bibitem{li2024crossfuse}
H.~Li and X.-J. Wu, ``Crossfuse: A novel cross attention mechanism based infrared and visible image fusion approach,'' \emph{Information Fusion}, vol. 103, p. 102147, 2024.

\bibitem{gu2023mamba}
A.~Gu and T.~Dao, ``Mamba: Linear-time sequence modeling with selective state spaces,'' \emph{arXiv preprint arXiv:2312.00752}, 2023.

\bibitem{lu2024structured}
C.~Lu, Y.~Schroecker, A.~Gu, E.~Parisotto, J.~Foerster, S.~Singh, and F.~Behbahani, ``Structured state space models for in-context reinforcement learning,'' \emph{Advances in Neural Information Processing Systems}, vol.~36, 2024.

\bibitem{ruan2024vm}
J.~Ruan and S.~Xiang, ``Vm-unet: Vision mamba unet for medical image segmentation,'' \emph{arXiv preprint arXiv:2402.02491}, 2024.

\bibitem{nguyen2022s4nd}
E.~Nguyen, K.~Goel, A.~Gu, G.~Downs, P.~Shah, T.~Dao, S.~Baccus, and C.~R{\'e}, ``S4nd: Modeling images and videos as multidimensional signals with state spaces,'' \emph{Advances in neural information processing systems}, vol.~35, pp. 2846--2861, 2022.

\bibitem{gu2020hippo}
A.~Gu, T.~Dao, S.~Ermon, A.~Rudra, and C.~R{\'e}, ``Hippo: Recurrent memory with optimal polynomial projections,'' \emph{Advances in neural information processing systems}, vol.~33, pp. 1474--1487, 2020.

\bibitem{tridao2024mamba}
A.~G. Tri~Dao, ``Transformers are ssms: Generalized models and efficient algorithms with structured state space duality,'' in \emph{International Conference on Machine Learning}, 2024.

\bibitem{liu2024vmamba}
Y.~Liu, Y.~Tian, Y.~Zhao, H.~Yu, L.~Xie, Y.~Wang, Q.~Ye, and Y.~Liu, ``Vmamba: Visual state space model,'' in \emph{International Conference on Machine Learning}, 2024.

\bibitem{cao2024novel}
Z.~Cao, X.~Wu, L.-J. Deng, and Y.~Zhong, ``A novel state space model with local enhancement and state sharing for image fusion,'' \emph{arXiv preprint arXiv:2404.09293}, 2024.

\bibitem{tang2022piafusion}
L.~Tang, J.~Yuan, H.~Zhang, X.~Jiang, and J.~Ma, ``Piafusion: A progressive infrared and visible image fusion network based on illumination aware,'' \emph{Information Fusion}, vol.~83, pp. 79--92, 2022.

\bibitem{xu2020aaai}
H.~Xu, J.~Ma, Z.~Le, J.~Jiang, and X.~Guo, ``Fusiondn: A unified densely connected network for image fusion,'' in \emph{Proceedings of the Thirty-Fourth AAAI Conference on Artificial Intelligence (AAAI)}, 2020, pp. 12\,484--12\,491.

\bibitem{zhao2023ddfm}
Z.~Zhao, H.~Bai, Y.~Zhu, J.~Zhang, S.~Xu, Y.~Zhang, K.~Zhang, D.~Meng, R.~Timofte, and L.~Van~Gool, ``Ddfm: denoising diffusion model for multi-modality image fusion,'' in \emph{Proceedings of the IEEE/CVF International Conference on Computer Vision}, 2023, pp. 8082--8093.

\bibitem{xie2023semantics}
H.~Xie, Y.~Zhang, J.~Qiu, X.~Zhai, X.~Liu, Y.~Yang, S.~Zhao, Y.~Luo, and J.~Zhong, ``Semantics lead all: Towards unified image registration and fusion from a semantic perspective,'' \emph{Information Fusion}, p. 101835, 2023.

\bibitem{Wang_2023_IF}
W.~Di, L.~Jinyuan, R.~Liu, and F.~Xin, ``An interactively reinforced paradigm for joint infrared-visible image fusion and saliency object detection,'' in \emph{Information Fusion}, 2023.

\bibitem{eskicioglu1995image}
A.~M. Eskicioglu and P.~S. Fisher, ``Image quality measures and their performance,'' \emph{IEEE Transactions on communications}, vol.~43, no.~12, pp. 2959--2965, 1995.

\bibitem{sheikh2006image}
H.~R. Sheikh and A.~C. Bovik, ``Image information and visual quality,'' \emph{IEEE Transactions on image processing}, vol.~15, no.~2, pp. 430--444, 2006.

\bibitem{piella2003new}
G.~Piella and H.~Heijmans, ``A new quality metric for image fusion,'' in \emph{Proceedings 2003 international conference on image processing (Cat. No. 03CH37429)}, vol.~3.\hskip 1em plus 0.5em minus 0.4em\relax IEEE, 2003, pp. III--173.

\bibitem{chen2024topiq}
C.~Chen, J.~Mo, J.~Hou, H.~Wu, L.~Liao, W.~Sun, Q.~Yan, and W.~Lin, ``Topiq: A top-down approach from semantics to distortions for image quality assessment,'' \emph{IEEE Transactions on Image Processing}, 2024.

\bibitem{ke2021musiq}
J.~Ke, Q.~Wang, Y.~Wang, P.~Milanfar, and F.~Yang, ``Musiq: Multi-scale image quality transformer,'' in \emph{Proceedings of the IEEE/CVF international conference on computer vision}, 2021, pp. 5148--5157.

\bibitem{zamir2022restormer}
S.~W. Zamir, A.~Arora, S.~Khan, M.~Hayat, F.~S. Khan, and M.-H. Yang, ``Restormer: Efficient transformer for high-resolution image restoration,'' in \emph{Proceedings of the IEEE/CVF conference on computer vision and pattern recognition}, 2022, pp. 5728--5739.

\end{thebibliography}

\begin{IEEEbiography}
[{\includegraphics[width=1in,height=1.25in,clip,keepaspectratio]{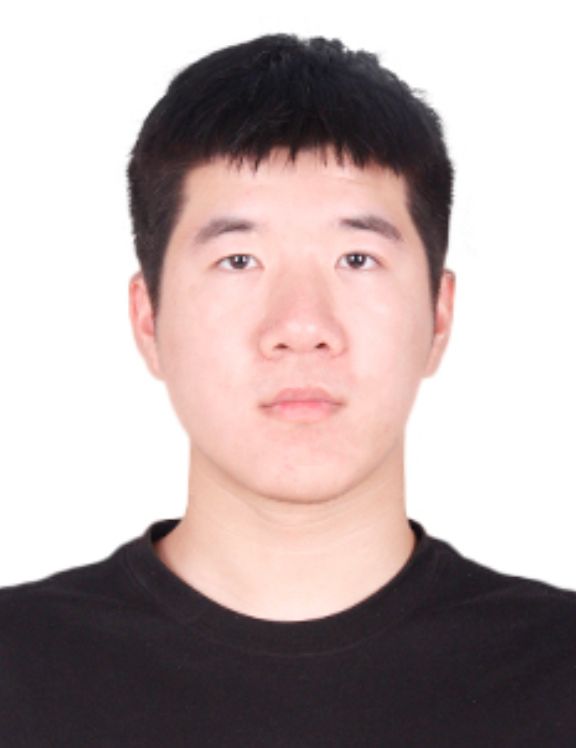}}]{Haolong Ma} received his bachelor's degree from the School of Computer Science at Heilongjiang Institute of Technology, China. He is currently a master's student at the Jiangsu Provincial Engineering Laboratory of Pattern Recognition and Computational Intelligence, Jiangnan University. His research interests include image fusion, natural language processing, and machine learning.

\end{IEEEbiography}

\begin{IEEEbiography}
[{\includegraphics[width=1in,height=1.25in,clip,keepaspectratio]{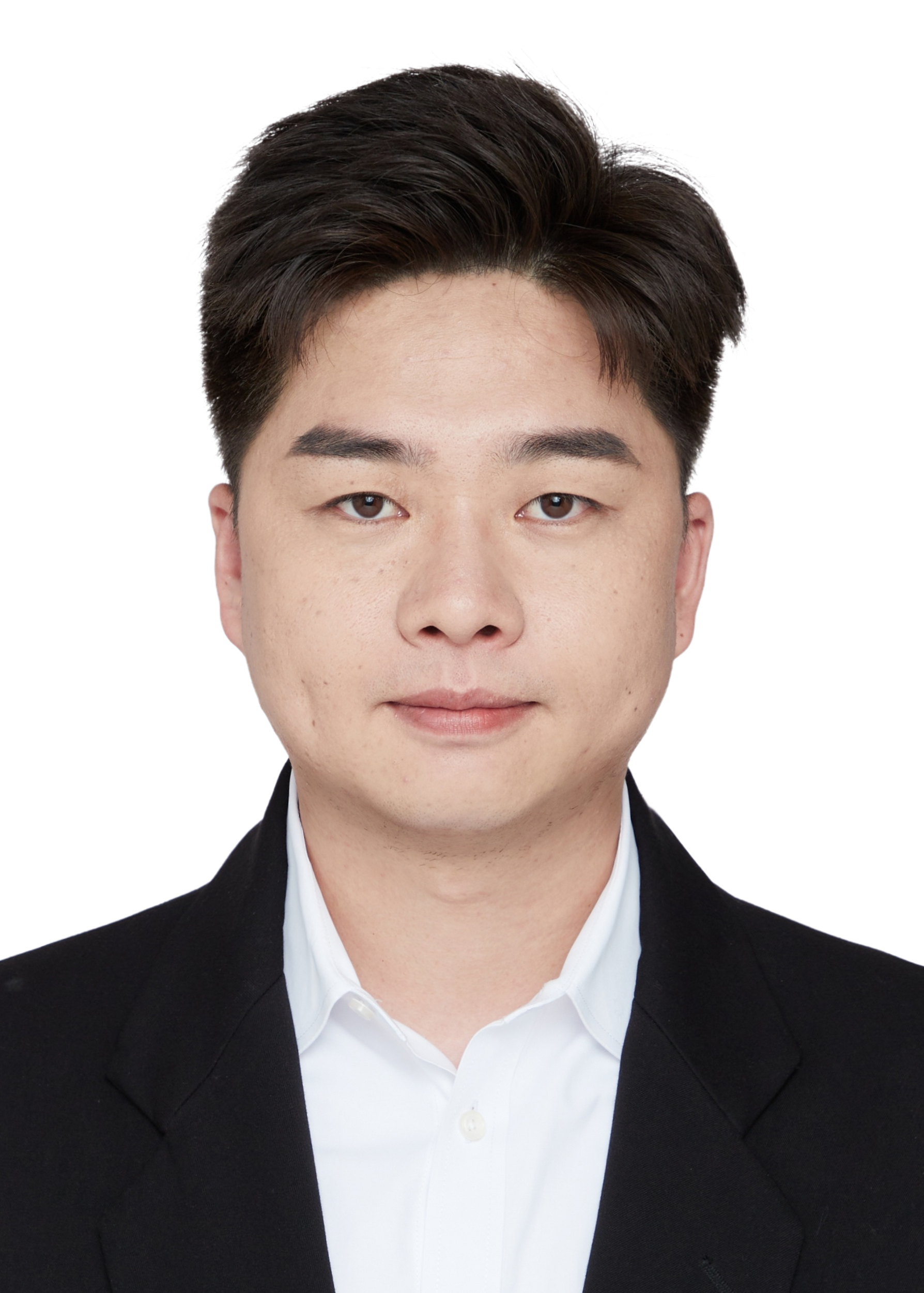}}]{Hui Li} (Member, IEEE) received the B.Sc. degree in School of Internet of Things Engineering from Jiangnan University, China, in 2015. He received the PhD degree at the School of Internet of Things Engineering, Jiangnan University, Wuxi, China, in 2022. He is currently a Lecturer at the School of Artificial Intelligence and Computer Science, Jiangnan University, Wuxi, China. His research interests include image fusion and multi-modal visual information processing. He has been chosen among the World's Top 2\% Scientists ranking in the single recent year dataset published by Stanford University (2021, 2022 and 2023).

He has published several scientific papers, including IEEE TPAMI, IEEE TIP, Information Fusion etc. He achieved top tracking performance in several competitions, including the VOT2020 RGBT challenge (ECCV20) and Anti-UAV challenge (ICCV21).

\end{IEEEbiography}

\begin{IEEEbiography}
[{\includegraphics[width=1in,height=1.25in,clip,keepaspectratio]{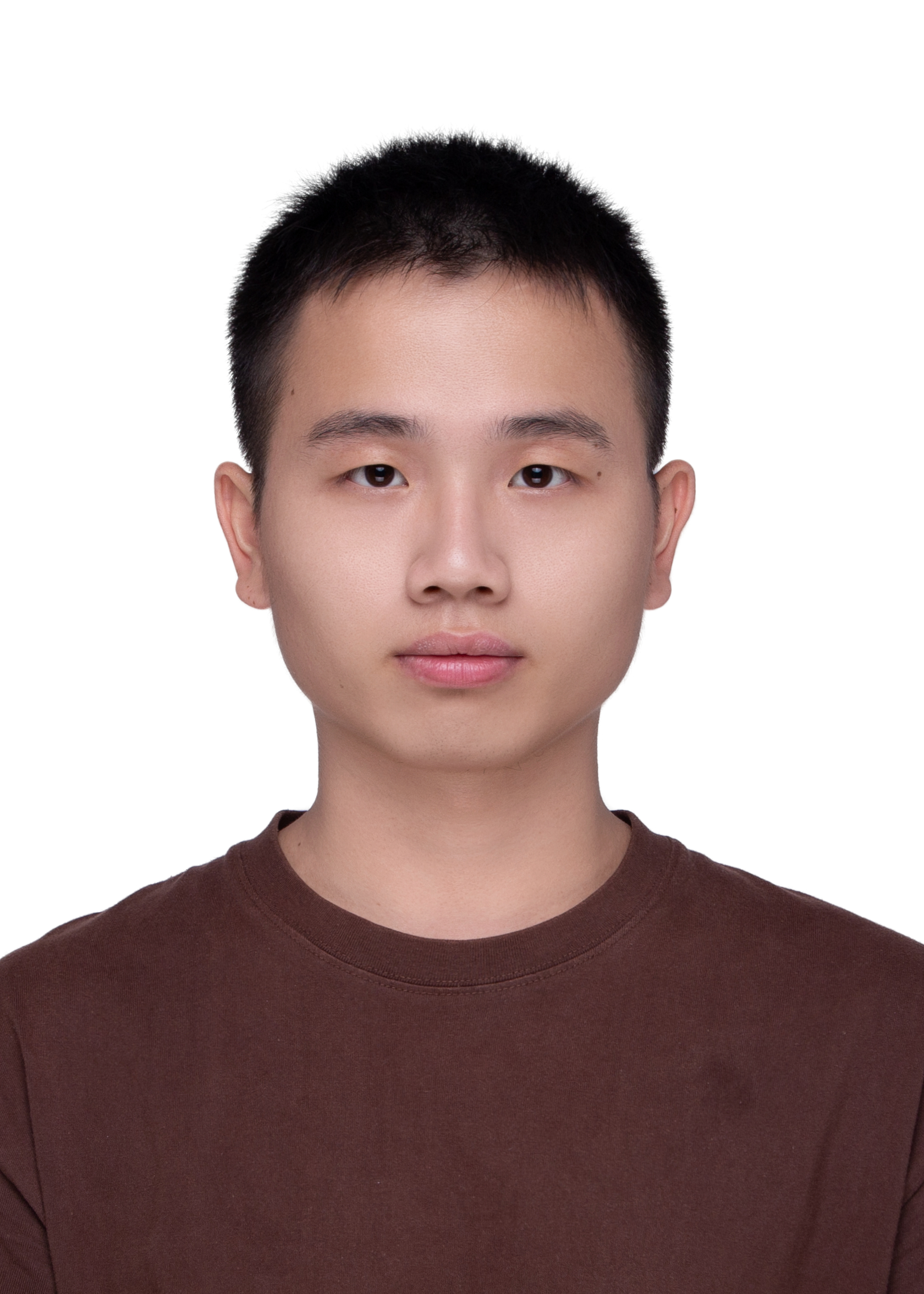}}]{Chunyang Cheng} is working toward the Ph.D degree at Jiangsu Provincial Engineering Laboratory of Pattern Recognition and Computational Intelligence, Jiangnan University, Wuxi, China. His research interests include image fusion and deep learning. He has published several scientific papers, including IJCV, Information Fusion, IEEE TIM etc.

\end{IEEEbiography}

\begin{IEEEbiography}
[{\includegraphics[width=1in,height=1.25in,clip,keepaspectratio]{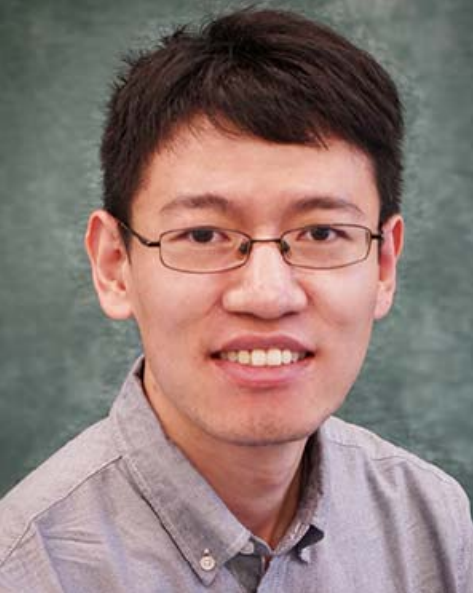}}]{Gaoang Wang} (Member, IEEE) Gaoang Wang  received the B.S. degree from Fudan University, Shanghai, China, in 2013, the M.S. degree from the University of Wisconsin-Madison, Madison, WI, USA, in 2015, and the Ph.D. degree from the Information Processing Laboratory, Electrical and Computer Engineering Department, University of Washington, Seattle, WA, USA, in 2019. In 2020, he joined the International Campus of Zhejiang University, Hangzhou, China, as an Assistant Professor. He is also an Adjunct Assistant Professor with the University of Illinois Urbana-Champaign Champaign, Champaign, IL, USA. In 2019, he joined Megvii US Office as a Research Scientist working on multi-frame fusion. In 2019, he then joined Wyze Labs working on deep neural network design for edge-cloud collaboration. 

He has authored or coauthored papers in many renowned journals and conferences, including IEEE TRANSACTIONS ON IMAGE
PROCESSING, IEEE TRANSACTIONS ON MULTIMEDIA, IEEE TRANSACTIONS ON
CIRCUITS AND SYSTEMS FOR VIDEO TECHNOLOGY, IEEE TRANSACTIONS ON
VEHICULAR TECHNOLOGY, Conference on Computer Vision and Pattern Recognition, International Conference on Computer Vision, European Conference on
Computer Vision, ACM Multimedia, and International Joint Conference on Artificial Intelligence. His research interests include computer vision, machine learning, artificial intelligence, multi-object tracking, representation learning, and active learning.

\end{IEEEbiography}

\begin{IEEEbiography}
[{\includegraphics[width=1in,height=1.25in,clip,keepaspectratio]{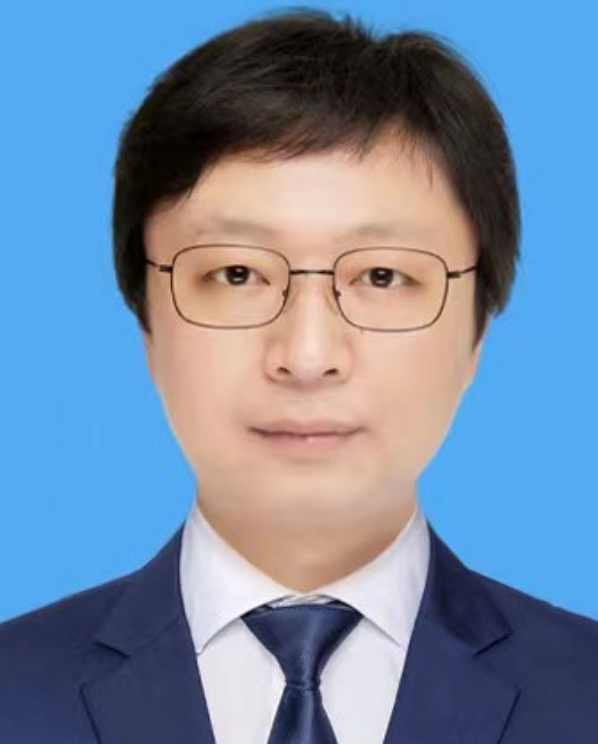}}]{Xiaoning Song} (Member, IEEE) received the B.Sc.
degree in computer science from Southeast University, Nanjing, China, in 1997, the M.Sc. degree in computer science from the Jiangsu University of
Science and Technology, Zhenjiang, China, in 2005,
and the Ph.D. degree in pattern recognition and
intelligence system from the Nanjing University of
Science and Technology, Nanjing, in 2010.

He was a Visiting Researcher with the Centre for
Vision, Speech, and Signal Processing, University of
Surrey, Guildford, U.K., from 2014 to 2015. He is
currently a Full Professor with the School of Artificial Intelligence and
Computer Science, Jiangnan University, Wuxi, China. His research interests
include pattern recognition, machine learning, and computer vision.

\end{IEEEbiography}

\begin{IEEEbiography}
[{\includegraphics[width=1in,height=1.25in,clip,keepaspectratio]{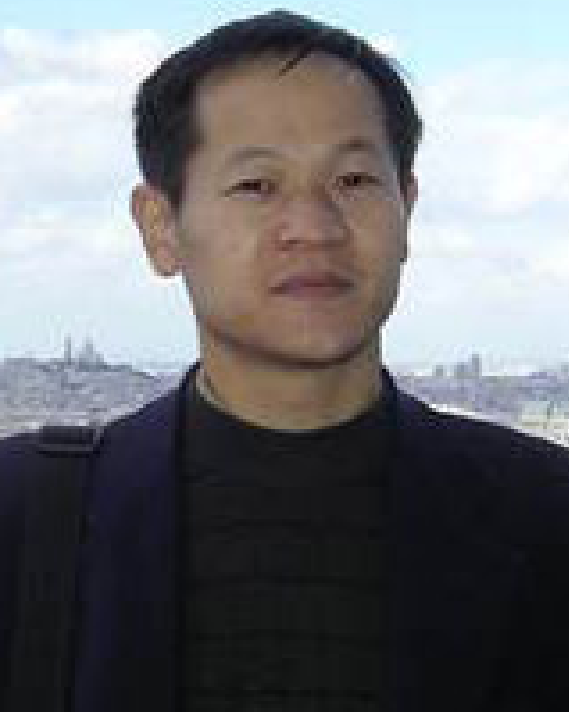}}]{Xiaojun Wu}  (Member, IEEE) received his B.S. degree in mathematics from Nanjing Normal University, Nanjing, PR China in 1991 and M.S. degree in 1996, and Ph.D. degree in Pattern Recognition and Intelligent System in 2002, both from Nanjing University of Science and Technology, Nanjing, PR China, respectively. He was a fellow of United Nations University, International Institute for Software Technology (UNU/IIST) from 1999 to 2000. From 1996 to 2006, he taught in the School of Electronics and Information, Jiangsu University of Science and Technology where he was an exceptionally promoted professor. He joined Jiangnan University in 2006 where he is currently a distinguished professor in the School of Artificial Intelligence and Computer Science, Jiangnan University. 

He won the most outstanding postgraduate award by Nanjing University of Science and Technology. He has published more than 400 papers in his fields of research. He was a visiting postdoctoral researcher in the Centre for Vision, Speech, and Signal Processing (CVSSP), University of Surrey, UK from 2003 to 2004, under the supervision of Professor Josef Kittler. His current research interests are pattern recognition, computer vision, fuzzy systems, and neural networks. He owned several domestic and international awards because of his research achievements. Currently, he is a Fellow of IAPR and AAIA.

\end{IEEEbiography}
\vfill

\end{document}